\documentclass[10pt,oneside]{article}

\usepackage  {graphics}
\usepackage[round]{natbib}
\usepackage{graphicx} 
\usepackage{verbatim}
\usepackage{epsfig}
\usepackage{latexsym}
\usepackage{longtable}
\usepackage{rotating}
\usepackage{wrapfig}
\usepackage{subfig}
\usepackage{amsmath,amssymb,latexsym,epsfig,color}
\usepackage{float}
\usepackage{amsthm}
\usepackage{makecell}
\usepackage{comment}
\usepackage{footnote}
\usepackage{lineno}
\usepackage{authblk}
\usepackage[sc]{mathpazo}

\usepackage{hyperref}
\hypersetup{
	colorlinks,
	linkcolor={blue!50!blue},
	citecolor={blue!50!blue},
	urlcolor={blue!80!blue}
}

\usepackage{booktabs}
\usepackage{graphicx}
\usepackage{multirow}
\usepackage{multicol}
\usepackage[export]{adjustbox}
\usepackage{amssymb,amsfonts}
\usepackage{amsmath}

\usepackage{tikz}

\usepackage{pifont}
\usepackage{array}
\newcolumntype{L}[1]{>{\raggedright\let\newline\\\arraybackslash\hspace{0pt}}m{#1}}
\newcolumntype{C}[1]{>{\centering\let\newline\\\arraybackslash\hspace{0pt}}m{#1}}
\newcolumntype{R}[1]{>{\raggedleft\let\newline\\\arraybackslash\hspace{0pt}}m{#1}}
\usepackage[english]{babel}
\usepackage{bm}
\usepackage{fancybox}
\usepackage{bookmark}

\usepackage{amsthm,bbm,mathtools,mathrsfs}
\usepackage{caption}
\usepackage{url}
\usepackage{enumerate}
\usepackage{lineno}
\usepackage{enumitem}
\usepackage{optidef}
\usepackage{xcolor}
\usepackage{soul}
\usepackage{threeparttable}
\usepackage{color}
\usepackage{cases}
\usepackage{amsmath}
\newtheorem{remark}{Remark}

\usepackage{algorithm,algpseudocode}
\usepackage{tabularx,booktabs}
\newcolumntype{Y}{>{\centering\arraybackslash}X}
\usepackage[titletoc]{appendix}

\usepackage{lineno}
\definecolor{myblue}{rgb}{0.01, 0.28, 0.9}
\algdef{SE}[SUBALG]{Indent}{EndIndent}{}{\algorithmicend\ }%
\algtext*{Indent}
\algtext*{EndIndent}
\usepackage{scrextend}
\allowdisplaybreaks

\begin{document}
\title{Interpretable Machine Learning for Traffic Congestion Prediction: Unveiling the Impact of Different COVID-19 Periods}
\author[a]{Dan Zhu}
\author[a]{Chi Sin Ng}
\author[a]{Litian Xie}
\author[a,b]{Yang Liu\footnote{Corresponding author, Phone: 65-6516-2334; Email: ceelya@nus.edu.sg}}
\affil[a]{Department of Civil and Environmental Engineering, National University of Singapore, Singapore}
\affil[b]{Department of Industrial Systems Engineering and Management, National University of Singapore, Singapore}
\date{\vspace{-5ex}}
\maketitle 
\begin{abstract}
Traffic congestion prediction plays a crucial role in mitigating congestion. However, the COVID-19 pandemic and associated government control measures have significantly altered urban travel behavior, making the traffic congestion prediction more complex.  This study aims to predict traffic congestion in Alameda County in the San Francisco Bay Area, California, during pre-lockdown, lockdown, and post-lockdown periods. We incorporate three external categories of data, i.e., weather conditions, seasonality factors, and COVID-19-related variables, and use Recursive Feature Elimination with Cross-Validation to identify important features across different periods and avoid potential overfitting. Based upon this, multiple advanced machine learning (ML) models, including support vector regression (SVR), multiple linear regression (MLR), recurrent neural networks (RNN), and long short-term memory (LSTM) networks, are trained and optimized through extensive experimentation and parameter tuning. Since LSTM has more hyperparameters and is more sensitive to tuning than the other ML methods used, we employ an adaptive parameter selection approach to optimize its hyperparameters, enhancing model accuracy and efficiency, rather than manually tuning parameters for SVR and RNN. 
These models are evaluated using Normalized Root Mean Square Error. {\color{blue}The results indicate that the bidirectional LSTM (Bi-LSTM) consistently outperforms the other models across all COVID-19 periods.} This superior performance can be attributed to the Bi-LSTM's bidirectional architecture, which effectively captures temporal dependencies by analyzing data both forward and backward in time. To address the limited interpretability of ML methods and provide valuable insights, we apply the Integrated Gradients (IG) technique to interpret the best-performing and differentiable Bi-LSTM's predictions. {\color{blue}Our analysis reveals that the new COVID-19 cases have a negatively dominant influence on traffic congestion during the lockdown and post-lockdown periods. The observed traffic reduction can be explained by heightened public risk awareness,  voluntary reductions in travel, and compliance with government-imposed mobility restrictions.} We also apply SHapley Additive exPlanations to SVR, given that IG is not applicable to this model. The results indicate that in the post-pandemic period, people have become more cautious—high new hospitalization discourages travel, reducing traffic congestion, while high fuel prices do not deter a shift toward private vehicle use, leading to increased congestion.

\end{abstract}
\section{Introduction}
\label{sec:intro}
Traffic congestion, as a significant global issue, has led to widespread negative consequences such as increased accident rates, unreliable travel times, elevated fuel consumption, and heightened air pollution. As reported in the 2019 Urban Mobility Report, traffic congestion in the United States caused an additional 8.8 billion hours of travel time and consumed an additional 3.3 billion gallons of fuel each year, leading to an estimated economic loss of approximately \$179 billion \citep{schrank2019urban}. Therefore, a traffic prediction model is needed to take timely preventive actions to avoid it.

Predicting traffic congestion is challenging due to its nonlinear, time-varying nature and the influence of various uncertain factors, such as weather and seasonal changes. Moreover, the COVID-19 pandemic, which began in late 2019, has significantly impacted urban mobility and should be considered when predicting traffic congestion. As of March 31, 2020, the COVID-19 pandemic had spread rapidly to 206 countries, with 750890 confirmed patients and 36405 deaths worldwide\footnote{https://www.who.int/docs/default-source/coronaviruse/situation-reports/20200331-sitrep-71-covid-19.pdf?sfvrsn=4360e92b\_8}. To avoid further spread of the COVID-19 pandemic,  many countries enforced strict lockdown and movement restrictions throughout 2020, drastically reducing mobility and altering traffic patterns \citep{dingil2021influence}. Because of it, 
restrictions began to ease in 2021 \citep{das2022impact}. However, the pandemic caused a lasting and irreversible impact on travel behavior. For example, many people switched to private transport modes such as cars, electric scooters and bicycles, avoiding public transportation\footnote{https://www.tomtom.com/en-gb/traffic-index/}. A survey conducted in Shanghai revealed that 41.38\% of the respondents switched to telecommuting, while 82.31\% of the public transport users transitioned to private modes during the work resumption period.
\cite{schrank2023urban} found that although traffic congestion in 2022 has nearly returned to 2019 levels, the distribution of congestion throughout the day has changed, reflecting the enduring effects of the pandemic on urban mobility. Despite the abundance of traffic congestion prediction models \citep{shaygan2022traffic}, most fail to incorporate COVID-19-related factors.

To effectively predict and compare nonrecurrent traffic congestion before, during and after the COVID-19 pandemic, this study selects Alameda County in the San Francisco Bay Area, California, as a case study due to its unique traffic dynamics during the pandemic, the availability of comprehensive traffic data and its high-density work environment. To curb the spread of the COVID-19 pandemic, the Alameda County government implemented several restriction measures in a phased approach as follows:
\begin{itemize}
    \item On 16 March 2020, Alameda County issued a Shelter-in-Place Health Order, directing residents to shelter at home starting 17 March \footnote{https://covid-19.acgov.org/covid19-assets/docs/response/hcsa-weekly-update-2020.05.18.pdf}. The order restricted activity, travel, and business functions to only the most essential needs. 
    \item On 19 June 2020, Alameda County issued a new health directive permitting the reopening of various activities, including outdoor museums, outdoor dining in restaurants, limited religious and cultural gatherings, indoor and outdoor retail operations (including malls) and outdoor fitness sessions.
\end{itemize}

Based on this, this study develops a model to predict traffic congestion in Alameda County, with a focus on how factors such as weather, seasonal variations, and COVID-19 impacts influence daily traffic congestion. Although previous studies have investigated congestion during evacuations triggered by natural disasters or severe weather events, such as hurricanes and heavy rainstorms \citep{wolshon2009temporospatial, li2015effects}, they cannot be directly applied to pandemic scenarios. This is because natural disasters and pandemics differ significantly in terms of impact duration. Although the effects of natural disasters are relatively short—usually lasting a few weeks or months—pandemics, like COVID-19, can persist for several months or even years \citep{li2021urban}. The unprecedented nature of a pandemic, coupled with the influence of multiple factors, makes traffic pattern prediction particularly complex and challenging. Given the capability of machine learning (ML) methods to handle complex and nonlinear data, as well as their strong performance in prediction tasks \citep{karlaftis2011statistical, liu2021deeptsp}, this study applies Support Vector Regression (SVR), Multiple Linear Regression (MLR), Recurrent Neural Networks (RNN), and Long Short-Term Memory (LSTM) Networks (a variant of RNN) to predict traffic congestion in Alameda County before, during, and after the COVID-19 pandemic.  Additionally,  to address the limited interpretability of these ML models, we use Integrated Gradients (IG),  proposed by \cite{sundararajan2017axiomatic},  to interpret complex neural networks by attributing the contribution of each input feature to the model's prediction. Considering that IG can only be utilized for differentiable models such as RNN and LSTM models, we apply another SHapley Additive exPlanations (SHAP) method, which was proposed by \cite{lundberg2017unified} based on game theory to analyze the feature importance of SVR's output. By leveraging these interpretability techniques, we gain deep insight into how various factors shape traffic congestion patterns across different pandemic periods, thereby providing a more transparent and explainable foundation for data-driven transportation planning and policy making.


\section{Literature Review}
\label{sec:liter}
This section reviews the literature closely related to this study, which can be divided into two categories: ML models for traffic congestion prediction and the impact of the COVID-19 pandemic on traffic patterns across different regions.
\subsection{Machine Learning Models in Traffic Congestion Prediction}
In the early stages of the traffic congestion prediction research, statistical methods were commonly used. These included time series approaches like auto-regressive integrated moving average (ARIMA) models \citep{yu2004switching}, Kalman filter techniques \citep{xie2007short}, and models based on hidden Markov processes \citep{qiao2018hybrid}. However, many of these approaches rely on assumptions of linearity and stationarity, which are not well suited to the complex and irregular nature of real-world traffic patterns. To address this limitation and improve prediction accuracy, researchers began adopting ML models to handle the nonlinear, dynamic, and spatiotemporal nature of traffic data. Among these, multiple linear regression (MLR) is the most common type of regression analysis for the purpose of prediction. \cite{lee2015prediction} proposed an MLR model which used 48 weather forecasting factors and six dummy variables to predict the traffic congestion on the roads near Ocean Beach in July and August 2014.
Besides, Support Vector Regression (SVR) has gained popularity due to its superior performance in traffic prediction compared to other methods \citep{deshpande2017performance}. For example, \cite{xu2014spatio} proposed a spatial-temporal variable selection-based SVR model, using high-dimensional traffic data from all available road segments for short-term urban traffic flow prediction. Similarly, \cite{yasir2022traffic} applied SVR to predict traffic congestion one week in advance using traffic and weather data from the previous week. In the case study in New Delhi, they found that the SVR model achieved high prediction accuracy, with an average RMSE of 1.12 for week-ahead congestion forecasts.

Given that traffic information is chronological, many studies have also employed recurrent neural networks (RNN) for traffic congestion prediction. \cite{huang2020traffic} developed a traffic congestion prediction system that integrates image analysis techniques with an RNN architecture. However, traditional RNNs struggle with long-term dependencies, as they tend to forget past data features over time. To address this, a Long Short-Term Memory (LSTM) model, a variant of RNN, has been widely leveraged for resolving RNN's limitations \citep{ma2015long}. \cite{shin2020prediction} proposed an LSTM-based traffic congestion prediction approach that corrects missing temporal and spatial values, and found that LSTM outperforms traditional RNN based on a mean absolute percentage error evaluation metric. Some studies also focused on enhancing the interpretability of deep learning models. For instance, \cite{wang2019traffic} introduced a bi-directional, path-based LSTM model tailored for forecasting speeds within urban networks. Based on this, they explained the output of the hidden layer through visualization and qualitative analysis. Besides, \cite{zarbakhsh2022predicting} calculated the SHAP values for their random forest model to explain the importance of influential features.  
\subsection{Impact of COVID-19 on Traffic Patterns across Different Regions}
Due to the sudden and unexpected outbreak of the COVID-19 pandemic, many mobility-related activities were restricted or suspended as part of preventive measures. Numerous studies attempted to compare traffic demand levels in 2020 with pre-pandemic levels (\emph{i.e.}, before December 2019) from different regions around the world. \cite{lee2020relationship} examined the relationship between traffic trends and the spread of COVID-19 in South Korea by using statistical methods to analyze data after the confirmation of the country's first COVID-19 case. Their findings indicated that a decrease in newly confirmed cases corresponds to an increase in traffic levels. \cite{ghanim2022ann} developed an artificial neural network (ANN) model to predict daily traffic demand at key junctions in the City of Doha, Qatar, incorporating seasonal and response measures during the pandemic. Similarly, \cite{zarbakhsh2022predicting} explored the impact of COVID-19 on traffic in Dublin, Ireland, by using four categories of urban data—mobility-based features, environmental factors, COVID-19-related metrics, and time-related variables—to predict the TomTom Traffic Index (which measures additional time spent on the road due to congestion). They applied several ML models, including random forests, SVR, and light gradient boosting machines, and found that mobility-based features were especially significant in their predictions. To investigate traffic congestion patterns in Shanghai, China, \cite{li2021urban} collected three months of data on road traffic speed from 446 Traffic Analysis Zones using Baidu Maps. They developed the singular value decomposition algorithm to extract key features of the road transport system and identified three main spatial-temporal change patterns influenced by the pandemic. Besides, \cite{yao2022understanding} first identified travelers' frequent trip chains and then analyzed changes in travel patterns using license plate recognition data over different periods, from the onset of COVID-19 to the post-pandemic era.

In summary, while there are fundamental studies that use ML models to predict traffic congestion, the majority do not account for COVID-19-related features. Even among those who consider the impact of the pandemic, most focus on broader traffic patterns and mobility trends rather than specific congestion prediction. One notable exception is \cite{zarbakhsh2022predicting}'s study, which incorporates COVID-19-related features to predict traffic congestion. However, they do not consider RNN-based methods, which are well-suited for capturing temporal dependencies in traffic data. To our knowledge, this study makes the first attempt to apply traditional RNN models and their variations for traffic congestion prediction in a different region, Alameda County.  Different from \cite{zarbakhsh2022predicting} who incorporated mobility-based features, we use external factors, including weather, seasonal, and COVID-19-related factors to understand how broad environmental conditions influence traffic congestion. More importantly, we further enhance the interpretability of our models by applying IG and SHAP methods to comprehensively understand the relationship between COVID-19-related features and daily traffic congestion prediction.
\subsection{Contributions and Organization}
This study endeavors to bridge critical gaps identified in existing literature, such as forecasting traffic demand amidst extraordinary circumstances, enhancing the interpretability of predictive methodologies, and specifically addressing the nuances of traffic prediction in the context of the COVID-19 pandemic. The contributions of the paper can be summarized as follows:
\begin{itemize}
    \item Firstly, this research investigates changes in traffic congestion across different pandemic periods: pre-lockdown, lockdown, and post-lockdown in Alameda County. Various data types, including weather conditions, seasonal factors, and COVID-19 pandemic indicators, are well integrated to assess their influence on daily traffic congestion levels. To effectively identify the most relevant features for our analysis, Recursive Feature Elimination with Cross-Validation (RFECV) is employed for feature selection across the three pandemic periods. This method iteratively removes the least important features and evaluates model performance using cross-validation, ensuring robustness and generalizability. One significant advantage of RFECV is its ability to automatically determine the optimal number of features, which eliminates the need for manual tuning and reduces the risk of overfitting. We respectively present the cross-validation score trends and feature importance rankings for each period to clearly illustrate how the features were selected. Additionally, we validate the effectiveness of the RFECV-based feature selection by comparing the performance of ML models with and without the inclusion of the significant ones. A clear drop in performance upon their removal underscores the importance of retaining all features selected by RFECV in the machine learning model.
    \item Secondly, we adopt a range of ML models to predict daily traffic congestion, including traditional approaches, such as SVR   as well as  MLR, and advanced deep learning methods like RNN and Long Short-Term Memory (LSTM). These deep learning models are specifically designed to capture long-range temporal dependencies in sequential data and are well-suited for time-series forecasting tasks. In particular, LSTM addresses the vanishing gradient problem commonly encountered in standard RNNs, enabling more effective learning from longer sequences. It is worth noticing that the proposed modeling framework is generalizable and can be readily applied to other regions with similar data characteristics. Considering the LSTM model has more parameters and is more sensitive to tuning than others, we deploy an adaptive parameter selection method instead of manual tuning for SVR and RNN to enhance the accuracy, efficiency, and adaptability of the LSTM model. To estimate and compare how well ML models can predict the target value, we adopt the Normalized Root Mean
    Square Error (NRMSE) as an error metric. {\color{blue}Our results show that the Bi-LSTM model consistently outperforms SVR and traditional neural network models for different periods based on error criterion NRMSE. } This can be attributed to its ability to ``reverse-learn'' the data, which helps infer future traffic conditions by learning patterns backward for the same day. To further enhance the effectiveness of our ML models, we compare their performance with that of a statistical Seasonal AutoRegressive Integrated Moving Average (SARIMA) model. The results show that the ML models are more flexible and better suited to capturing the dynamics of traffic congestion, particularly during the lockdown and post-lockdown periods.
    \item  Finally, we apply the IG and SHAP methods to provide model interpretability and derive deep and valuable insights for Alameda County, which are applicable to regions with similar urban and traffic characteristics. {\color{blue}IG is adopted to identify the relationships between the input features and predicted values for the best-performing and differentiable Bi-LSTM model. The results show that new COVID-19 cases play a negatively dominant role during the lockdown and post-lockdown periods, which emphasizes the importance of COVID-19-related factors in traffic congestion prediction, and the reduced traffic is likely due to heightened public concern, voluntary avoidance behavior, or compliance with mobility restrictions.} Besides, during the post-lockdown period, new cases and new hospitalizations over the past week show a diminishing impact on traffic congestion for the prediction day. This lagged effect indicates that people may not immediately adjust their travel behavior in response to rising cases or hospitalizations, possibly due to fatigue, changing risk perceptions, delayed awareness, or gradual policy implementation. To evaluate the robustness of the IG results, we tested three different baselines, including zero, mean, and past-week average. The attribution maps derived under these baselines display consistent patterns, reinforcing the reliability of the insights drawn from IG. To complement the IG analysis, we also apply the SHAP method to explain feature importance in the SVR model, as IG is not applicable to SVR. We find that weekend, day of the week, and COVID-related variables are the most influential factors affecting the daily traffic congestion across different pandemic periods. Notably, weekends consistently lead to lower congestion. Furthermore, in the post-lockdown phase, heightened caution leads people to stay home more frequently when hospitalization rates are high, resulting in lower traffic congestion. Besides, despite rising fuel prices, there is a clear trend toward increased private vehicle use, which in turn drives up traffic congestion.  These insights emphasize the need for pandemic-aware transport policies, such as dynamic congestion pricing, public transit incentives, and early-stage public messaging, to manage mobility effectively.
\end{itemize}



The subsequent sections of the paper are structured as follows. Section \ref{Mat&met} displays data collection and analysis, feature selection methods, and the machine learning models used in the study. Section \ref{Res} presents the performance evaluation of the models and explores model interpretability through the IG and SHAP methods. Finally, Section \ref{conclu} summarizes the main findings, traffic management insights, and potential future research directions.
\section{Materials and Methods}
\label{Mat&met}
In this section, we provide a comprehensive overview of the data collection process, followed by an analysis of the collected data. We then present the feature selection techniques used to refine the dataset. Subsequently, we introduce the ML models for predicting traffic congestion and interpret these models through the IG and SHAP methods.
\subsection{Data Collection and Analysis}
Daily traffic congestion fluctuates due to various seasonal and non-seasonal factors, with the COVID-19 pandemic introducing additional, unprecedented variables. We begin by identifying the traffic congestion metrics in Alameda County. Then, we introduce three main categories of factors that may influence traffic congestion during the pandemic, including weather-related factors, seasonal variations, and COVID-19-related indicators.
\subsubsection{Traffic Congestion Index}
The Performance Measurement System (PeMS)\footnote{https://pems.dot.ca.gov/}, as an assessment framework in California State's freeway infrastructure, predominantly active in urban freeway corridors due to a high density of monitoring equipment, PeMS extensively tracks congestion levels. Vehicle detector stations are the fundamental data providers for PeMS, delivering real-time information on traffic flow and vehicle occupancy, which informs the calculation of numerous performance metrics and the generation of diverse analytical reports. 

\begin{figure}[ht]
    \centering
    \includegraphics[width=0.9\linewidth]{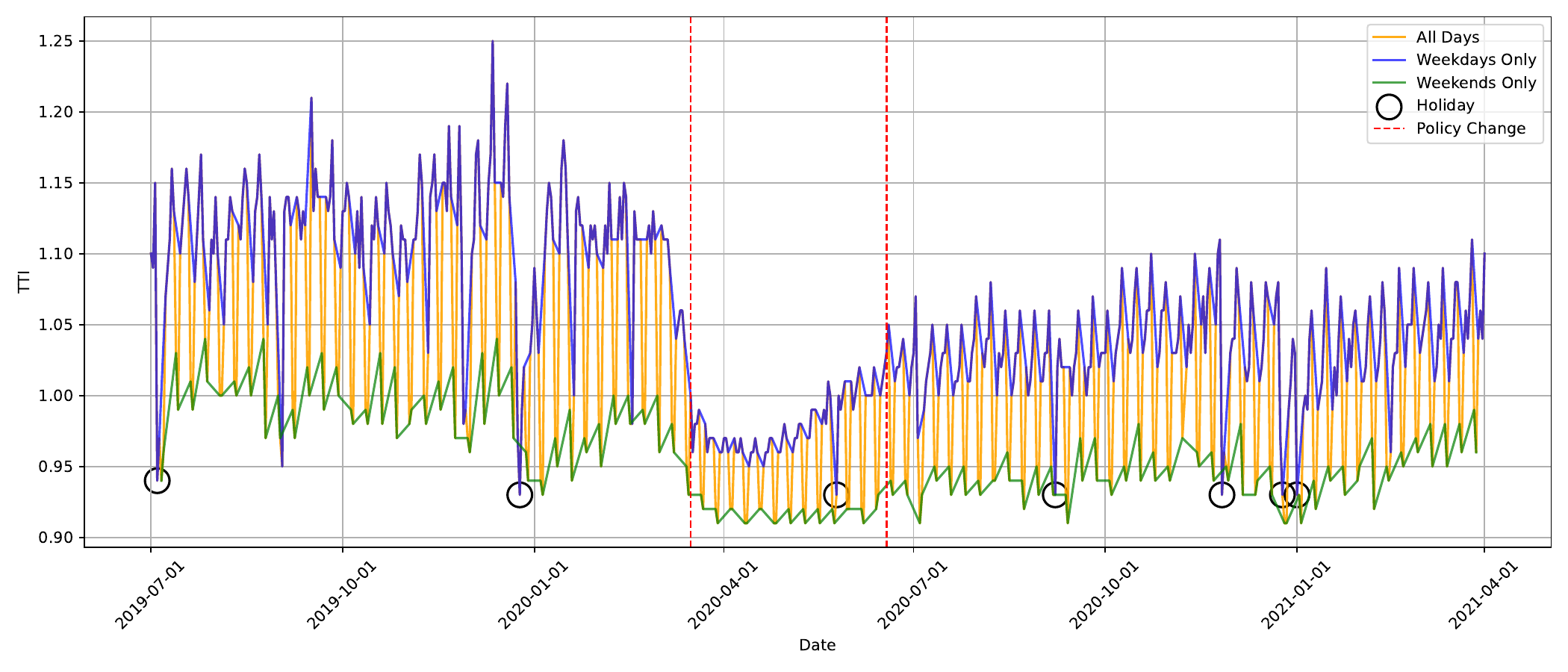}
    \caption{TTI levels across different periods} 
    \label{fig:TTI_total}
\end{figure}

Travel Time Index (TTI) is one of the performance measures calculated by PeMS and one of the common traffic congestion indicators \citep{Lokesh2019EvaluationOT}. TTI refers to the ratio of the average travel time for all vehicles across a freeway segment to the free-flow travel time. The free-flow travel time is the amount of time to traverse the freeway segment distance traveling at 60 miles per hour. A higher TTI indicates a higher congestion level as the travel time is longer and vice versa.  A detailed explanation of how TTI is calculated by PeMS is provided in Appendix A for reference.
Based on differences in restrictions imposed by  Alameda County, we define three periods: 1) July 1, 2019, to March 16, 2020, the pre-lockdown period, with unrestricted transportation;
2) March 17, 2020, to June 18, 2020, the lockdown period, with restrictions on transportation;
3) June 19, 2020, to March 31, 2021, the post-lockdown period, with gradual loosening of the transportation restriction. Figure \ref{fig:TTI_total} presents TTI variations across different pandemic periods. 
For the pre-lockdown period (Before the first red dashed line), the TTI shows considerable fluctuation but generally remains at higher levels. This variability suggests typical traffic patterns with working and non-working days, likely influenced by regular workday commutes and other socio-economic activities. Regarding the lockdown period (between the 2nd red dashed lines), there is a significant and abrupt drop in the TTI as the policy change (lockdown) comes into effect. The consistently low TTI values during this period reflect the substantial decrease in traffic congestion due to strict travel restrictions and stay-at-home orders. In terms of the post-lockdown period (after the 2nd red dashed line), the TTI gradually rises as restrictions begin to ease, indicating a rebound in traffic activity. However, the TTI does not return to pre-lockdown levels, which could suggest a lasting change in commuting patterns, possibly due to the adoption of remote work or ongoing social distancing measures.

\subsubsection{Seasonal Factors}
The first group of factors consists of seasonal influences. Under normal conditions, traffic congestion can fluctuate based on the type of day, the time of day, and the time of year. Congestion patterns also vary significantly during the holidays. In the San Francisco Bay Area, traffic is particularly sensitive to seasonal factors compared to other regions, largely due to the area's dynamic work culture, which is more prominent on weekdays than weekends or holidays. Therefore, we include seasonal factors as independent variables to capture these unique traffic patterns. Inputs such as day of the week, weekend, and holiday (\emph{e.g.}, New Year, Thanksgiving, Christmas) are incorporated to account for these variations. To validate that the traffic congestion level due to seasonality, the data points for the weekday and weekend were separated, and the trends were compared as well. Figure \ref{fig:TTI_total} displays the comparative trend analysis.

For the pre-lockdown period (Before the first red dashed line), the comparison reveals a stark contrast between weekdays and weekends. Weekdays exhibit higher TTI values, indicative of routine rush-hour congestion caused by regular work commutes and commercial activities. Weekends show a lower TTI, reflecting the reduced traffic that is typically seen on days when fewer people are traveling to work or school. 
In terms of the lockdown period (between the 2 red dashed lines),  the distinction between weekdays and weekends diminishes during this period. The implementation of lockdown measures leads to a dramatic drop in weekday traffic, which brings the TTI values closer to those observed on weekends. This suggests that the lockdown effectively neutralized the usual spike in traffic associated with weekdays, as non-essential work and travel were curtailed.
Regarding the post-lockdown period (after the second dashed line), as restrictions begin to lift, a gradual divergence in the TTI between weekdays and weekends becomes apparent. The weekday TTI starts to increase, indicating the resumption of work and social activities. However, the increase is modest compared to pre-lockdown levels, suggesting a possible shift in work patterns, such as the adoption of remote working. 

Additionally, the holiday feature is also important. Figure \ref{fig:TTI_total} highlights the days classified as public holidays. It is evident that TTI values drop noticeably on these circled days, underscoring the significant impact of holidays on traffic congestion. 

\subsubsection{Weather Factors}
The second group of factors contains weather factors. The weather data in this study were collected from the Weather Station ``Central Station'' located in Alameda. We extracted historical weather data from the website Weather Underground\footnote{wunderground.com}, a San-Francisco-based company that provides local and long-range weather forecasts, weather reports, maps, and tropical weather conditions for locations worldwide. The seasonal variables considered include humidity (\%), pressure (in), temperature (°F), wind speed (mph), and precipitation (in).
\subsubsection{COVID-19 Related Factors}
For the pandemic status, a set of indicators is published by the Alameda County Public Health Department. The numbers for the Alameda County Local Health Jurisdiction come from the state's communicable disease tracking database, CalREDIE (\emph{i.e.,} California Department of Public Health). These data are updated daily and available for public access. The assumption is that individuals may have voluntarily reduced their mobility and travel activities in response to the pandemic. The pandemic status indicators utilized in this study include new daily cases, 
new daily deaths, new daily hospitalizations, and the number of vaccinated people.
\subsection{Feature Selection Method}
Prior to the modeling process, feature selection is performed by eliminating redundant or irrelevant data that can lead to overfitting to enhance the model's performance. In this study, 
we apply Recursive Feature Elimination with Cross-Validation (RFECV) for its systematic and thorough approach to feature selection. Unlike basic feature elimination techniques that may rely on a single metric or a fixed dataset, RFECV integrates cross-validation, providing a more holistic view of a feature's importance across multiple data subsets \citep{mishra2022feature}. This process ensures that the selected features are both important and contribute to a stable and generalized model. RFECV is especially useful for datasets where the relationship between the features and the target variable is complex, as was the case with the traffic data during pandemic restrictions. One of the biggest advantages of this method is its ability to automatically determine the optimal number of parameters.

The RFECV process mainly includes three steps. The first step is to train a Random Forest Classifier (RFC) based on our dataset with all available features.{\color{blue}The first step is to train a \texttt{RandomForestRegressor (RFR)} based on our dataset using all available features.}  After training, the second step extracts the feature importance scores provided by the RFC. The final step conducts iterative feature elimination by evaluating whether removing the weakest feature (\emph{i.e.}, the feature with the lowest importance score) would improve the model's cross-validation accuracy until the removal of additional features no longer improve accuracy or the number of remaining features reaches a predefined minimum threshold. The final set of features selected through this process will be used in our subsequent analyses. By applying this procedure to each timeframe, we identify the final set of features presented in Table \ref{Tab:feature_select}. 
\begin{table*}[!ht]
\caption{Feature Selection Results Across Different Time Periods}
\centering
\resizebox{\linewidth}{!}{
\begin{tabular}{cccc}
\toprule
\makecell[c]{} & \makecell[c] {Period 1} & \makecell[c] {Period 2} & \makecell[c] {Period 3}\\ 
\midrule 
\makecell[c]{Optimal selected \\features \& number} & \makecell[c] {Holiday, weekend,\\ temperature, dew point,\\ humidity, wind speed,\\ precipitation, fuel price,\\
day of the week} & \makecell[c] {New cases, holiday,\\ weekend, temperature,\\wind speed, day of the week} & \makecell[c] {New cases, new hospitalizations, \\holiday, weekend, \\temperature, dew point, \\humidity, pressure,\\ wind speed, fuel prices, \\day of the week}\\
\bottomrule
\end{tabular}}
\label{Tab:feature_select}
\end{table*}

\begin{figure}
\centering
\subfloat[Cross validation score  \label{subfig:pre_cross}]{\includegraphics[width=6.5cm]{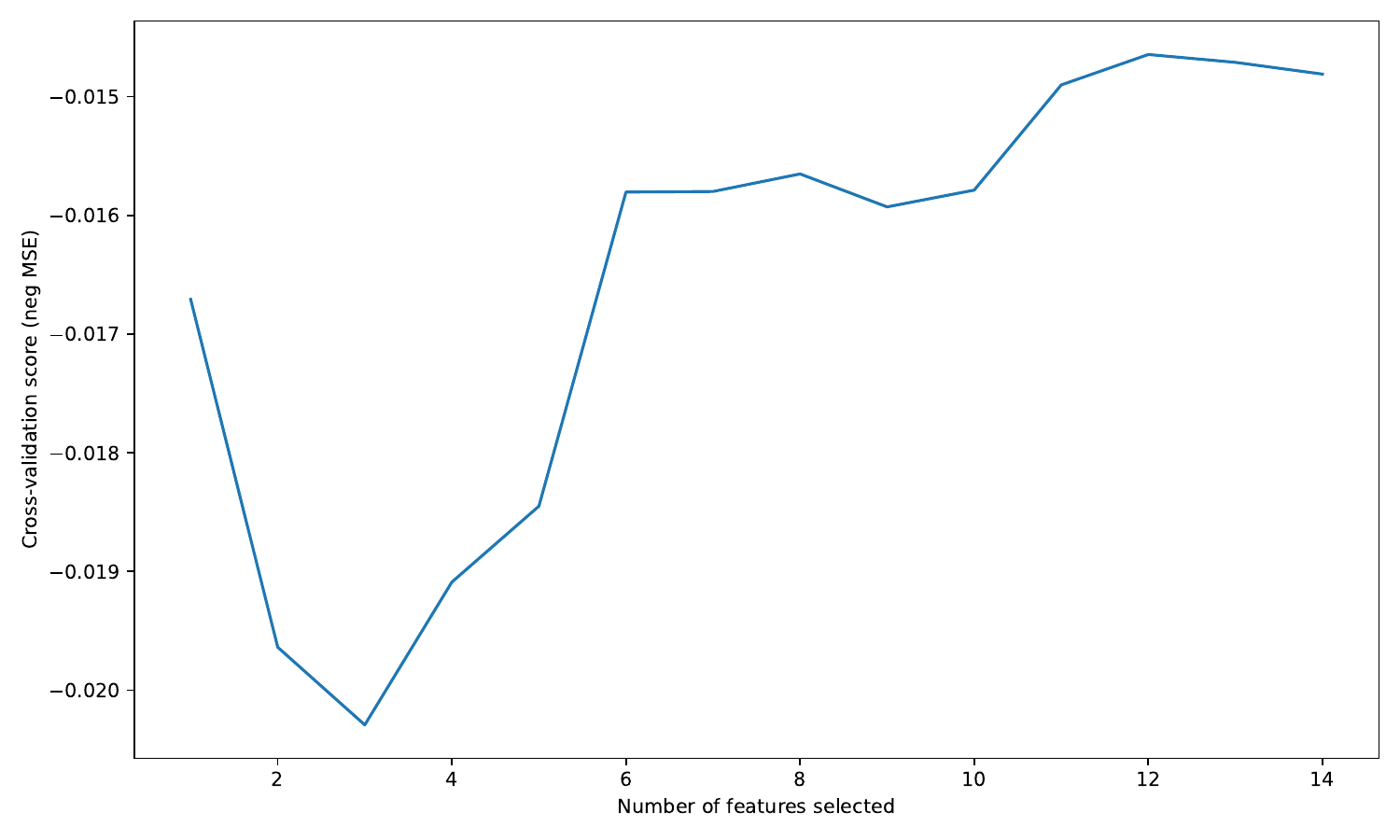}}
\subfloat[Feature importance \label{subfig:pre_import}]{\includegraphics[width=6.5cm]{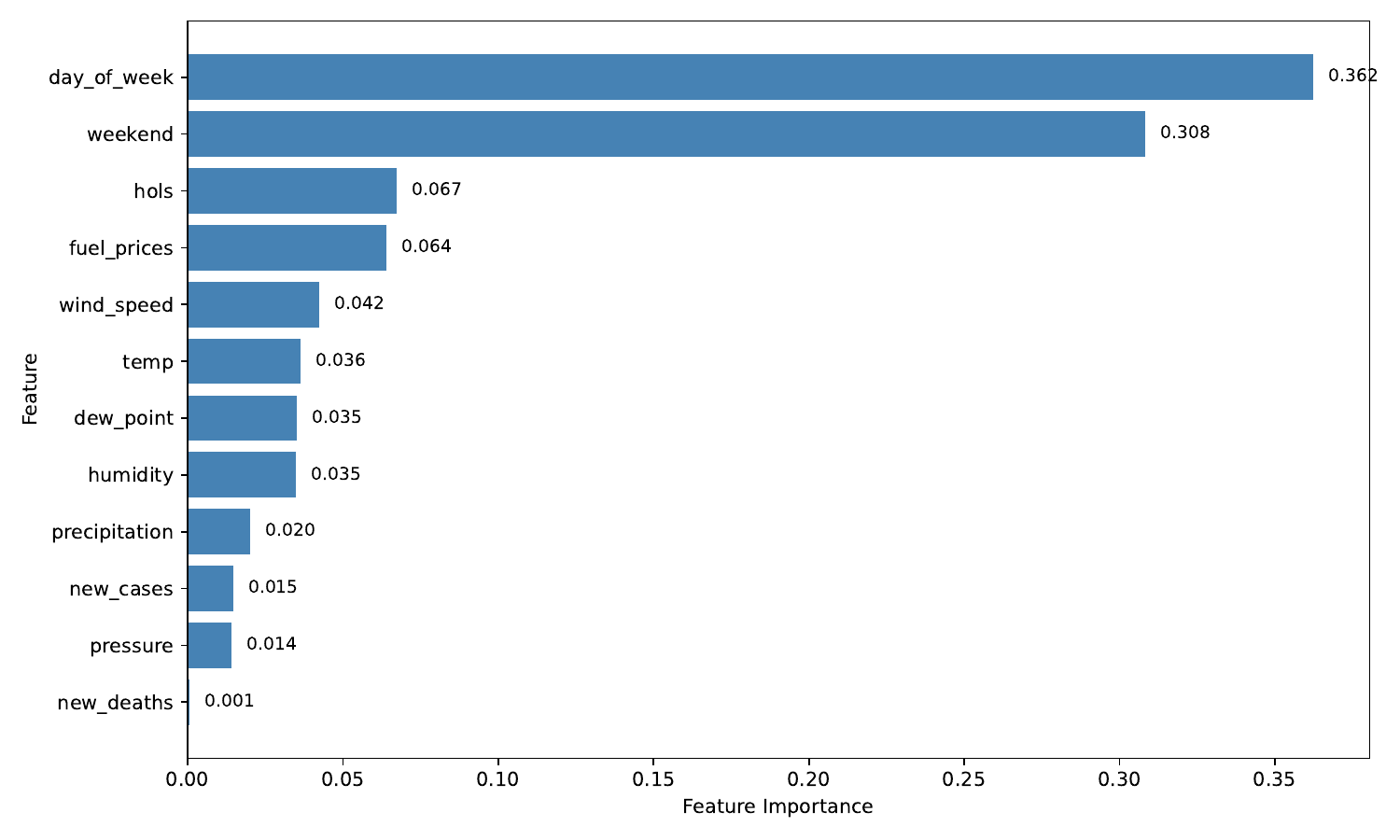}}\\
\caption{RFECV-based feature selection for the period 1}
\label{fig:REFCV1}
\end{figure}

\begin{figure}
\centering
\subfloat[Cross validation score \label{subfig:dur_cro}]{\includegraphics[width=6.5cm]{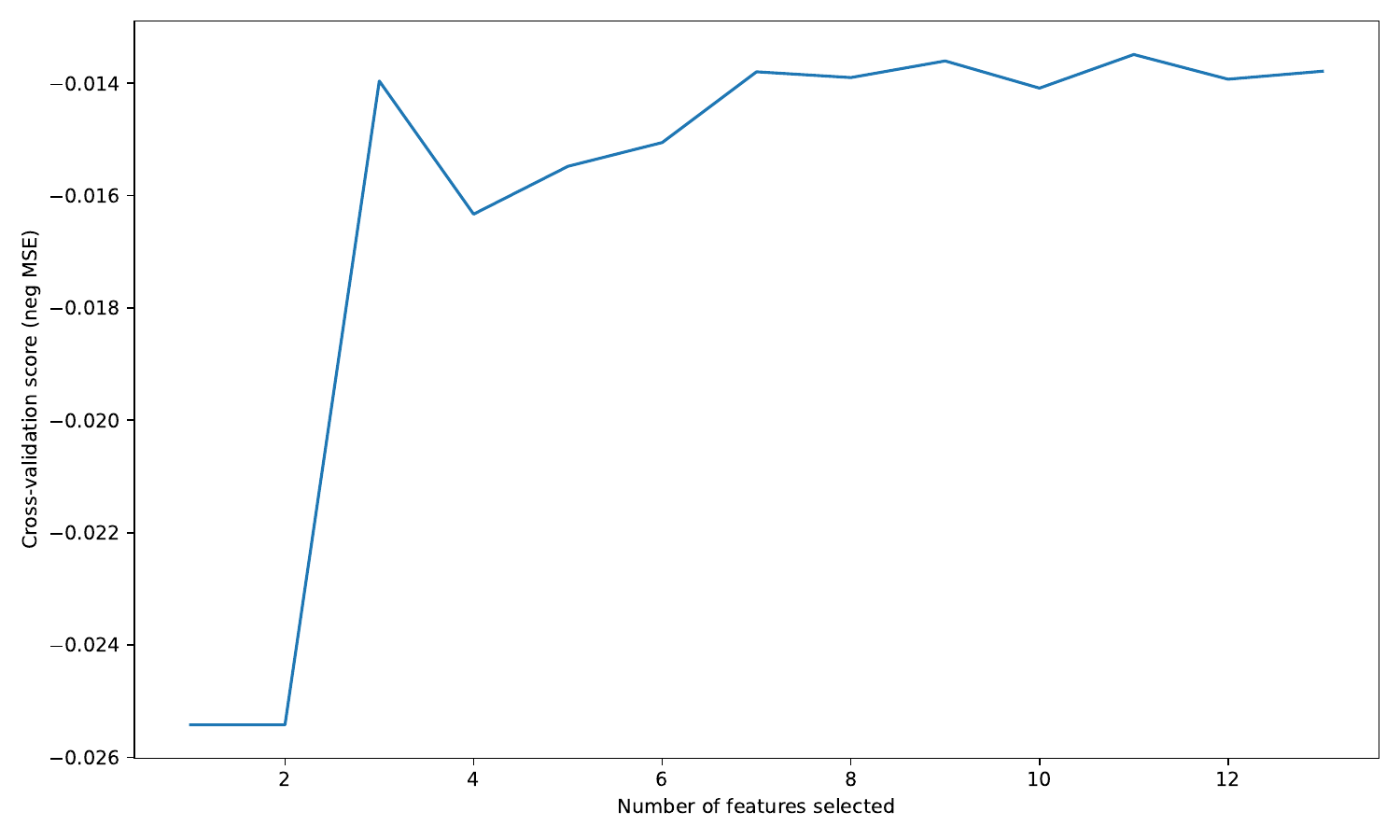}}
\subfloat[Feature importance  \label{subfig: dur_import}]{\includegraphics[width=6.5cm]{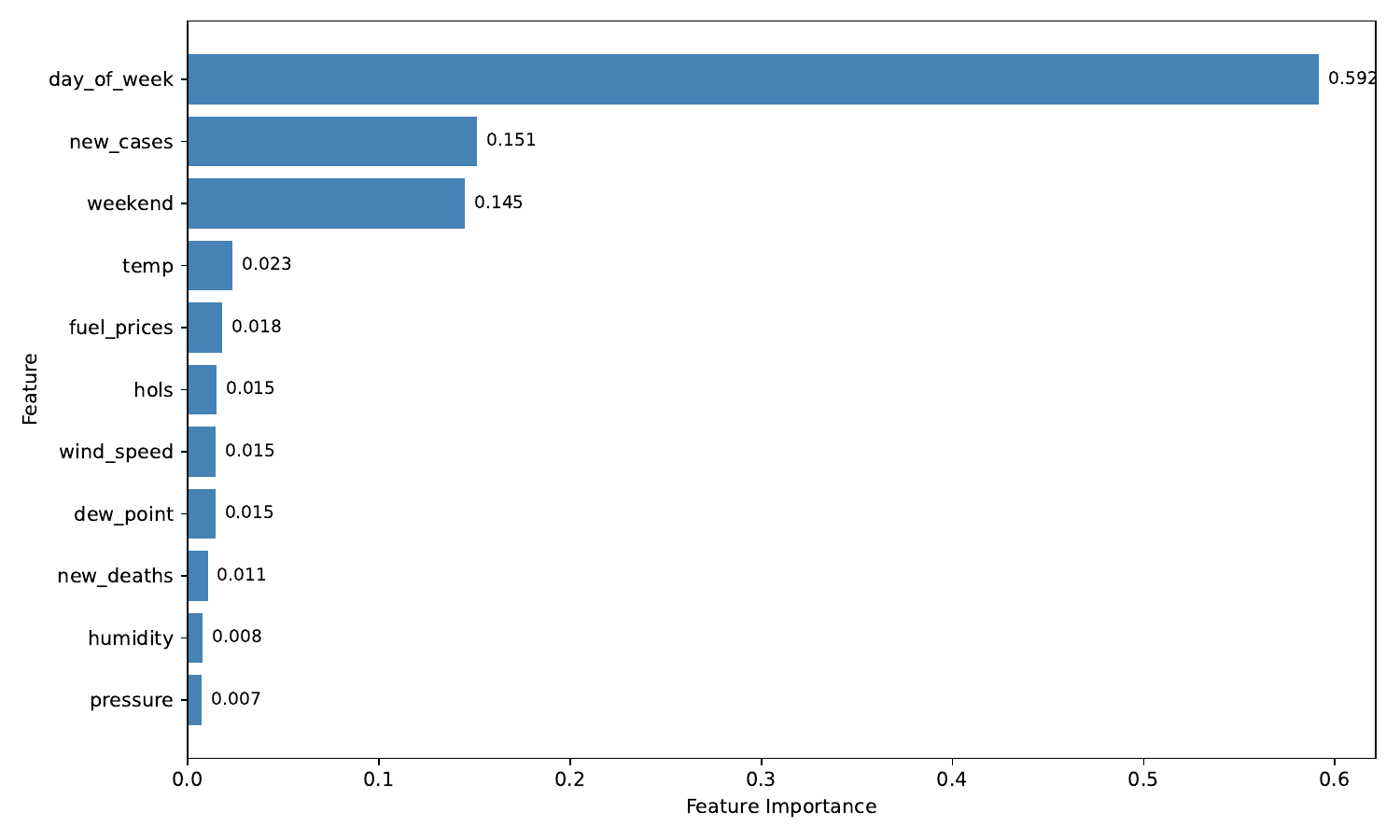}}\\
\caption{RFECV-based feature selection for the period 2}
\label{fig:REFCV2}
\end{figure}

\begin{figure}
\centering
\subfloat[Cross-validation score \label{subfig:aft_cro}]{\includegraphics[width=6.5cm]{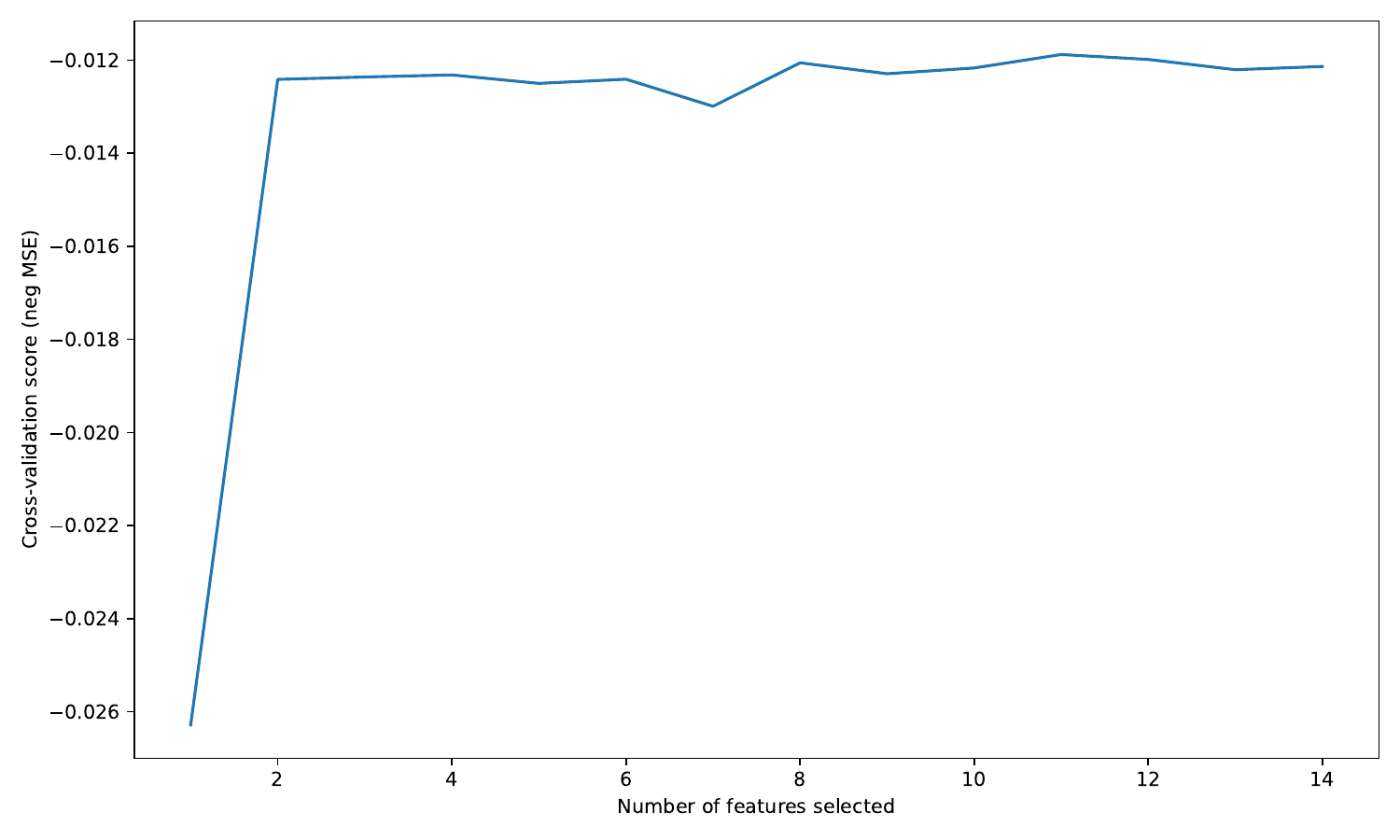}}
\subfloat[Feature importance \label{subfig:aft_import}]{\includegraphics[width=6.5cm]{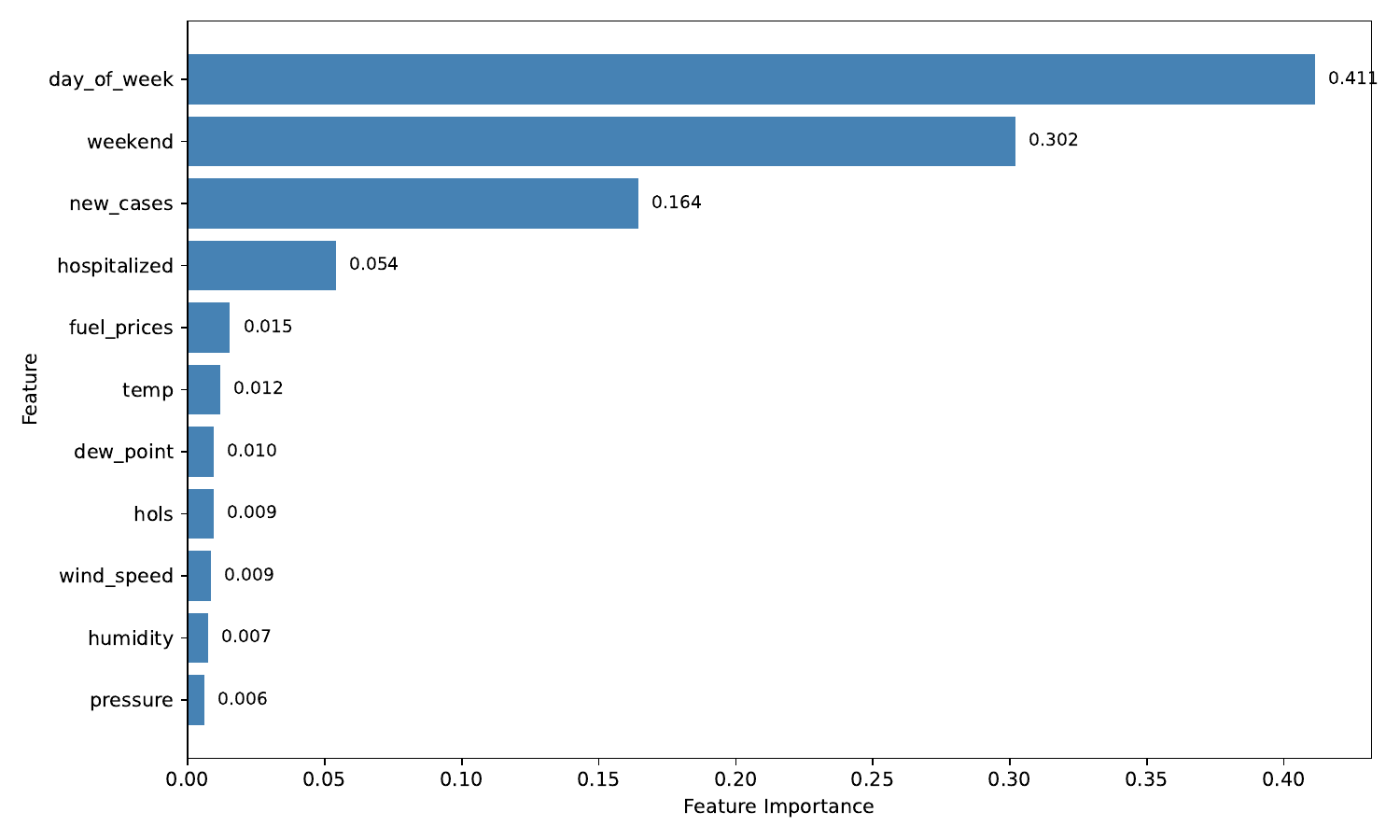}}\\
\caption{RFECV-based feature selection for the period 3}
\label{fig:REFCV3}
\end{figure}

To provide more details about the feature selection process, we draw the following three figures \eqref{fig:REFCV1}-\eqref{fig:REFCV3} to display CV scores and feature importance rankings corresponding to each period, respectively. 
\begin{itemize}
   \item \textbf{Pre-lockdown period}: The CV score curve in this period exhibits significant volatility as the number of features increases, with no clear minimum or plateau. This suggests a high sensitivity to the selected feature set and more complex, less structured mobility patterns. To maintain a balance between model stability and representational power, we select the top 10 features, which together cover over 95\% of cumulative feature importance. This choice mitigates overfitting risks without sacrificing essential information.

    \item \textbf{Lockdown period}: The CV score improves steadily with more features but plateaus after approximately six variables. This indicates a more stable travel behavior during the pandemic, likely shaped by consistent external restrictions (\emph{e.g.}, travel limitations posed by government). Therefore, we select the six features that offer near-optimal CV performance. In particular, we do not strictly follow the order of feature importance; instead, we prioritize the combinations that yield the best validation score, reflecting an adaptive and performance-driven approach.

    \item \textbf{Post-lockdown period}: A clear inflection point appears in the CV score curve around 10–11 features. As no degradation in performance is observed with 11 features, and all contribute significantly to the model, we retain all features for this period to ensure that the model captures potential rebound effects and evolving travel dynamics.
\end{itemize}
Moreover, the validity of our feature selection approach is supported by the actual model prediction performance. Consistently across all models, the pre-lockdown period exhibits the lowest prediction accuracy, which aligns with the observed volatility in its CV score curve and the more chaotic, unstructured travel patterns of that phase. In contrast, the lockdown and post-lockdown periods achieved significantly better predictive performance, which corresponds well to their smoother CV score trends and more structured mobility behavior. 

Note that the data are divided into three distinct periods, corresponding to the periods of the COVID-19 pandemic and the associated transportation restrictions. Period 1 contains 260 records, period 2 contains 94 records, and period 3 contains 286 records. For each period, the dataset is further divided into two subsets: training and testing datasets. {\color{blue} Specifically, the first 80\% of the records (in chronological order) are used as the training dataset, and the remaining 20\% are used as the testing dataset.} The training dataset comprises 80\% of the data. This subset is instrumental in the model learning process, enabling the discovery of intricate relationships between inputs and outputs. The testing dataset, comprising the remaining 20\% of the data, is reserved for model evaluation and serves as an independent set to assess the model's predictive accuracy. These records are vital for testing the model's performance in simulating real-world traffic conditions.
\begin{remark}
   Although more data are available for the pre-lockdown and post-lockdown periods, we limit their duration to a comparable range, given that only three months of data are available for the lockdown period. This approach ensures a fair comparison across different pandemic phases. We acknowledge that this constraint may reduce the overall volume of training data; however, the consistent performance of the ML models demonstrates their potential effectiveness even in data-scarce environments. Future research could explore the impact of larger and unbalanced datasets or apply transfer learning techniques to further enhance model generalizability.
\end{remark}
\subsection{Machine Learning Models}
After identifying the most influential factors through feature selection, this section applies three machine learning (ML) models, including support vector regression (SVR), multiple linear regression (MLR),  recurrent neural network (RNN), and long short-term memory (LSTM) networks to predict the traffic congestion level for three periods. Unlike traditional statistical approaches, ML models possess greater generalization power, allowing them to capture intricate patterns and adapt more effectively to the dynamic nature of traffic networks \citep{yang2020evaluation}. 
\subsubsection{Support Vector Regression}
Support Vector Regression (SVR) extends Support Vector Machines to regression tasks. It aims to find a function that approximates the relationship between the input features and the continuous output values while maintaining a specified margin of error, controlled by a parameter $\epsilon$. Unlike traditional regression, Support Vector Regression (SVR) focuses on minimizing model complexity and relies only on the data points outside the margin, known as support vectors. Its flexibility allows for the use of different kernel functions—such as `linear' (linear), `rbf' (radial basis function), or `poly' (polynomial)—depending on the linearity of the traffic data, making SVR effective in predicting continuous values across various applications. For our study, after the fine-tuning process, our SVR models are configured with `poly' for both periods 1 and 3, whereas period 2 gives better results with `linear'. Besides, proper data normalization is essential to avoid bias due to varying feature scales. In our experiment, we employ \texttt{MinMaxScaler} from \texttt{Scikit-learn} to scale features to a range $[0, 1]$  by subtracting the minimum value of the feature and then dividing by the range (maximum value - minimum value). This standardization ensures that the SVR model is not biased towards variables with large scales.

\subsubsection{Multiple Linear Regression}
Multiple Linear Regression (MLR) is one of the most fundamental and widely used statistical learning methods for modeling the relationship between a dependent variable and one or more independent variables \citep{tranmer2008multiple, noorossana2010statistical}. It assumes a linear relationship of the form:

\[
y = \beta_0 + \sum_{i=1}^{n} \beta_i x_i + \epsilon
\]

where \( y \) is the target variable, \( x_i \) are the input features, \( \beta_i \) are the model coefficients, and \( \epsilon \) is the error term. Despite its simplicity, MLR remains a strong baseline for many prediction tasks, offering clear interpretability and efficient computation. In our study, MLR is used as a benchmark model to compare with more complex machine learning methods. To ensure fair comparison, all input features are scaled to a [0, 1] range using \texttt{MinMaxScaler} from \textit{Scikit-learn}, which prevents variables with large numerical ranges from disproportionately influencing the regression coefficients. While MLR assumes linearity and independence among predictors, its ease of implementation and explainability make it a valuable tool for preliminary analysis and feature interpretation in traffic congestion prediction.

\subsubsection{Recurrent Neural Networks}
Recently, RNN, as a deep learning model, has become increasingly popular due to its strong applicability to capture temporal dependencies and patterns in time-series data, which makes it well-suited to our traffic prediction. Specifically,  RNN learns by updating a state at the current timestep depending on the prior output at the previous state and the input at the current timestep. Memory is realized by configuring each hidden neuron with a feedback loop that passes the current output as the input during the next step, which enables the network to learn temporal patterns and dependencies over time. Due to the RNN network's use of loops, backpropagation through time is utilized to train the network \citep{werbos1990backpropagation}. That is, the network's loss is propagated backward not only through the layers of the network but also through the sequence of time steps.
\subsubsection{Long Short-Term Memory Networks}
Traditional RNN is susceptible to issues such as vanishing or exploding gradients when training large networks, primarily due to the backpropagation process. This becomes particularly problematic when dealing with extended multivariate traffic time series. Specifically, increasing the number of hidden layers in the network results in more derivatives being multiplied with each pass. If these products are too large, the gradient may explode, causing arithmetic overload. Conversely, if the derivatives are very small, continuous multiplication can cause the gradient to vanish, potentially leading to arithmetic underflow \citep{shaygan2022traffic}. Hence, RNN can only deal with the problem of sequence handling for short-term dependencies but struggles with modeling long-term dependencies due to the vanishing gradient problem.

To capture potential long-term dependencies, we employ a long short-term memory (LSTM) network, a popular type of gated RNN \citep{schmidhuber1997long}. Unlike traditional RNN, LSTM networks use specialized memory units within their hidden layers to store information over extended time periods. Each memory unit is composed of cells that maintain a dynamic state through self-loops, enabling the model to better remember temporal sequences. The flow of information in each cell is regulated by three gates—input, output, and forget gates—which control the addition, retrieval, and deletion of information. This mechanism helps LSTMs retain relevant information over longer time spans, thereby mitigating the challenges associated with vanishing gradients. LSTM networks can be separated into unidirectional and bidirectional types. In a unidirectional LSTM (Uni-LSTM), the information flows only in one direction—typically from the past to the future within the sequence. A bidirectional LSTM (Bi-LSTM), on the other hand, processes the input sequence in both forward and backward directions. We will implement these two types of LSTM models in traffic congestion prediction. Note that in this study, a lookback window of 7 days is chosen for the Bi-LSTM model based on both domain knowledge and empirical considerations. This is because traffic congestion patterns often exhibit strong weekly cycles due to differences between weekdays and weekends. Using a 7-day window enables the model to effectively capture these temporal patterns while maintaining a manageable input size to avoid overfitting.

Considering that the LSTM model has more hyperparameters to tune than RNN, due to the introduction of three gates (\emph{i.e.}, input, forget, and output gates), manual hyperparameter tuning may be inefficient and impractical since we need to train LSTM models for periods 1, 2, 3, respectively. To address this, we adopt an adaptive parameter selection for LSTM by using \texttt{Keras\_Tuner}. This process systematically searches for the optimal combination of hyperparameters (\emph{e.g.}, number of units, activation function, dropout rate, and learning rate) to minimize validation loss. Specifically, we define a model-building function, which first initializes the Tuner using \texttt{Keras\_Tuner.RandomSearch} to explore the hyperparameter space, then 
searches for the best hyperparameters by training multiple models with different hyperparameter combinations to identify the optimal configuration, and finally builds and trains the final model using the best hyperparameters.

\section{Results}
\label{Res}
This section first analyzes and compares the performance of different ML prediction models, then applies IG and SHAP methods to provide the interpretability of ML models. Subsequently, we validate the impact of selected features on our ML models by removing significant ones. Finally, we discuss how our model can be well extended to incorporate spatial granularity.
\subsection{Model Evaluation}
We evaluate model performance by first comparing the predictive results of the five ML models and then benchmarking them against the traditional Seasonal Auto-Regressive Integrated Moving Average (SARIMA) model.
\subsubsection{Prediction Performance of Machine Learning Models}
\begin{figure}[ht]
    \centering
    \includegraphics[width=0.9\linewidth]{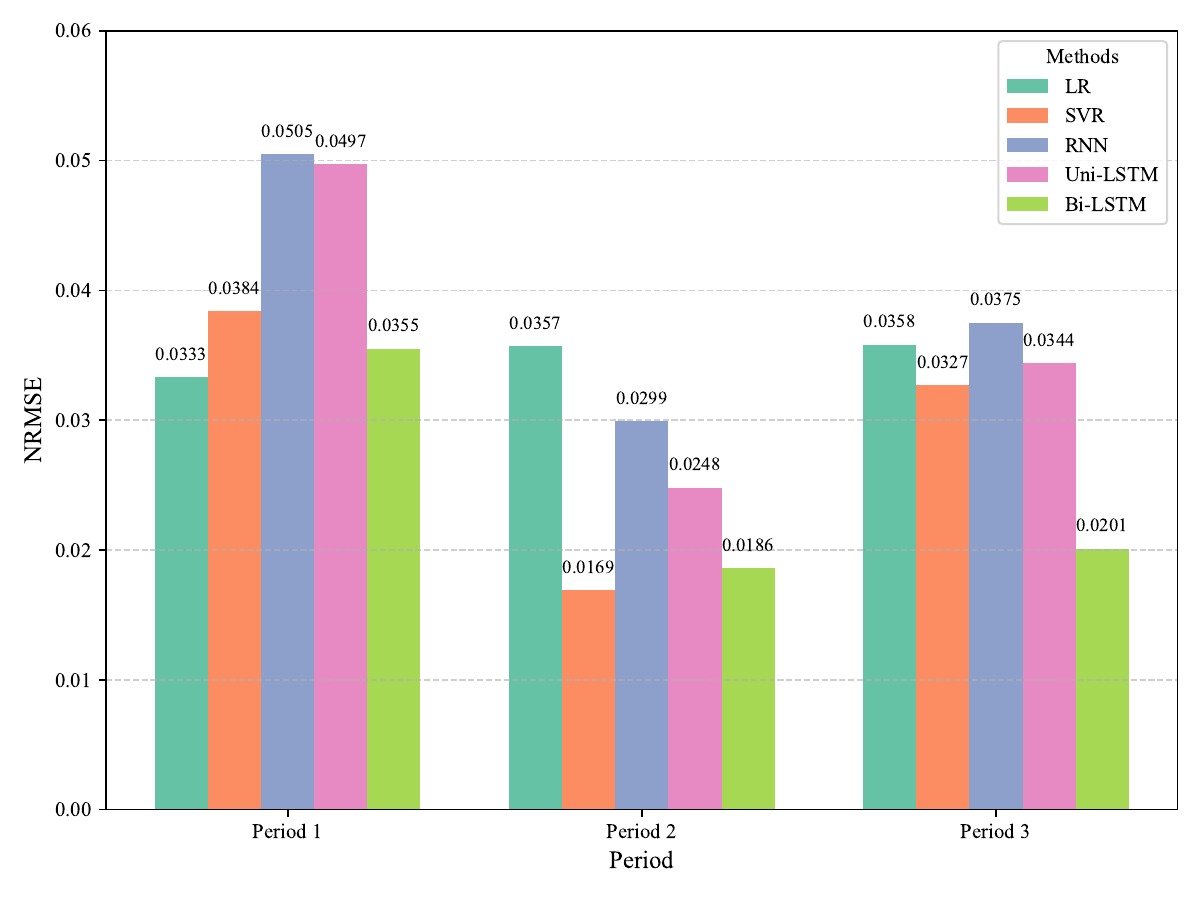}
    \caption{Comparison of ML models based on NRMSE across time periods}
    \label{fig:per_total}
\end{figure}

We compare the performance of the Bi-LSTM models with other models using Normalized Root Mean Square Error (NRMSE), a widely used metric for assessing prediction accuracy. NRMSE normalizes RMSE, which measures the average difference between actual and predicted values, divided by the mean of actual values. A lower NRMSE indicates a more accurate model. One key advantage of NRMSE is that it mitigates the risk of underestimating or overestimating errors that RMSE might introduce. Besides, by making error measurements scale independent, it enhances comparability across different datasets and reduces sensitivity to outliers. These properties make it particularly suitable for our models, as TTI values are on a small scale. We denote $y_i$ and $\hat{y}_i$ as the estimated and actual TTI values, respectively, and $\bar{y}$ as the mean actual TTI, given by $\sum_{i=1}^n y_i$, where $n$ represents the number of test data points. The mathematical formulation of NRMSE is as follows:
\[
NRMSE = \frac{\sqrt{\frac{1}{n}\sum_{i=1}^{n}\left(y_i-\hat{y}_i\right)^2}}{\bar{y}}.
\]
Figure \ref{fig:per_total} shows the NRMSE values of the traffic prediction models developed for the three periods. Across all periods, the Bi-LSTM model almost achieves the lowest NRMSE values, which confirms the efficacy of the Bi-LSTM model in traffic prediction. This can be attributed to its capability to capture temporal relationships more comprehensively by analyzing data from the past to the future and the future to the past.

\begin{figure}
\centering
\subfloat[SVR\label{subfig:SVM-1}]{\includegraphics[width=5.5cm]{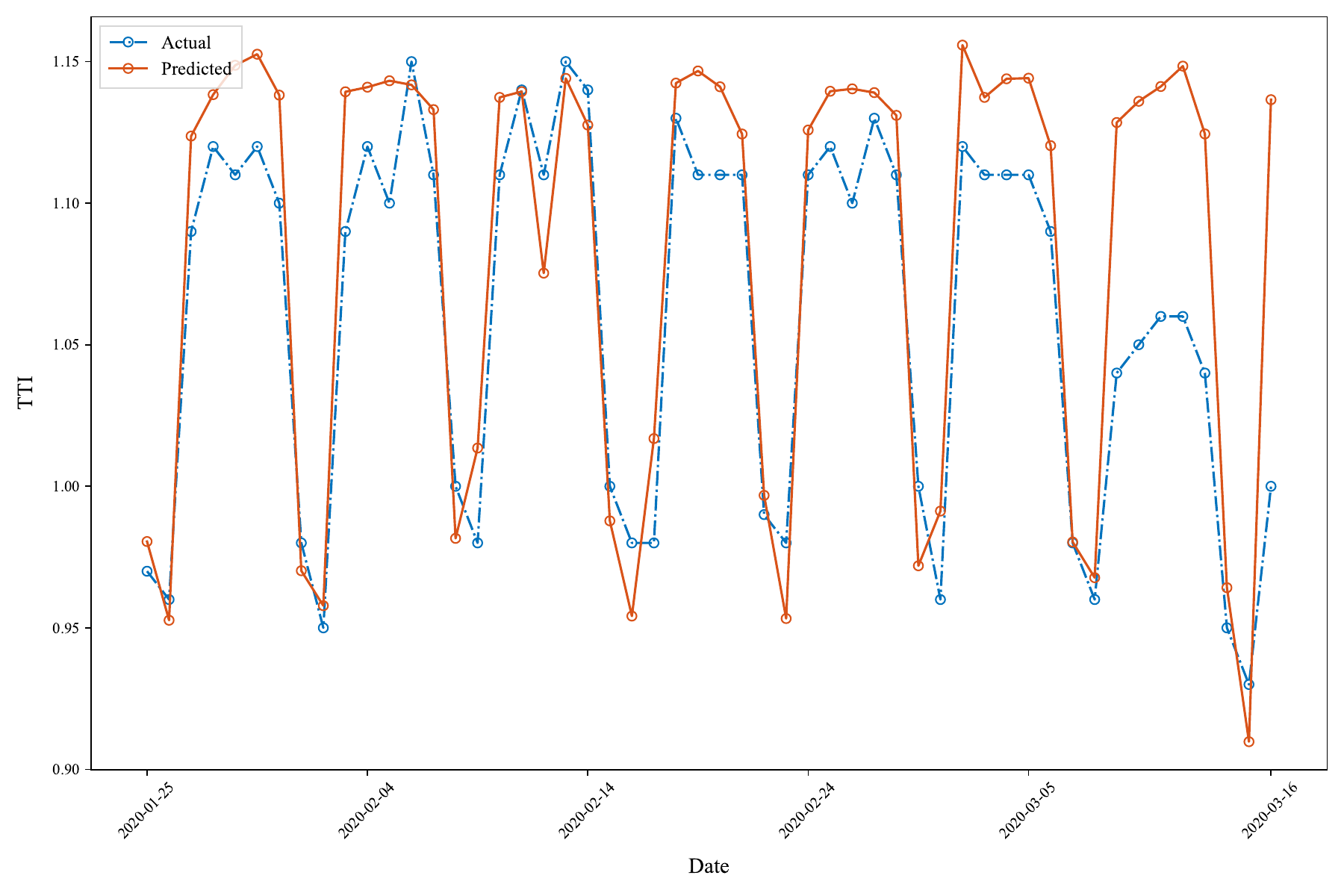}}
\subfloat[MLR\label{subfig:LR-1}]{\includegraphics[width=5.5cm]{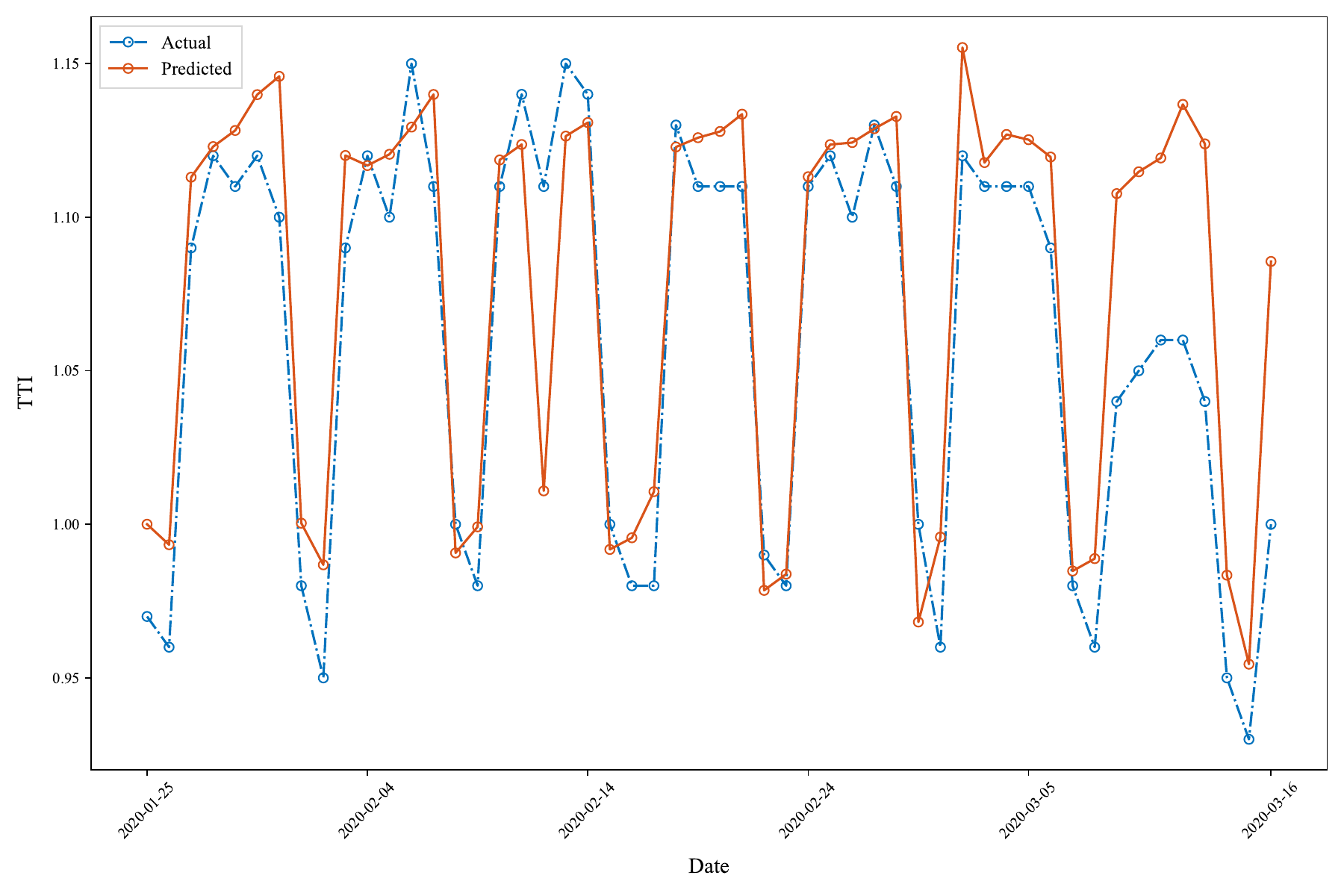}}
\subfloat[RNN\label{subfig: RNN-1}]{\includegraphics[width=5.5cm]{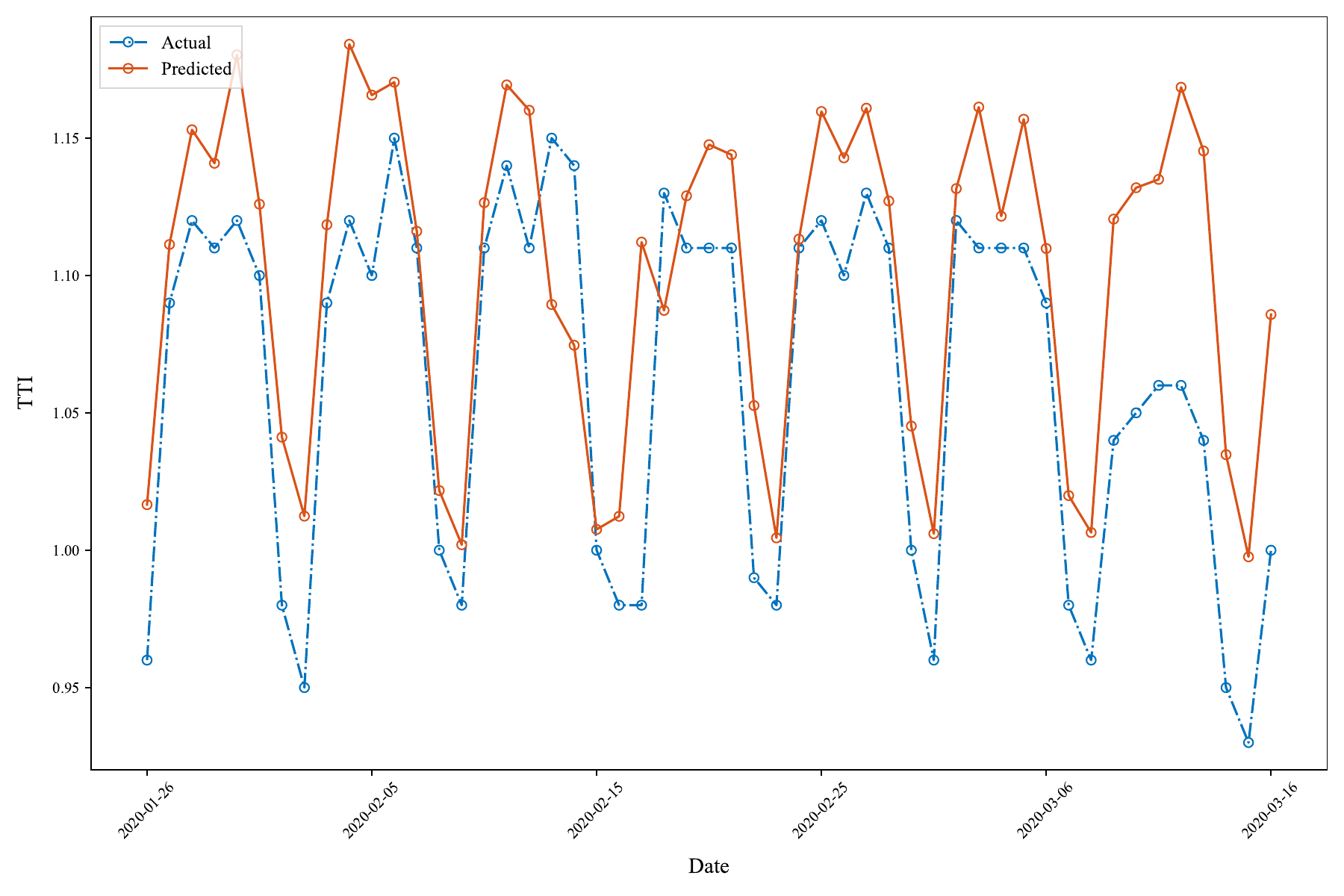}}\\
\subfloat[Uni-LSTM\label{subfig:Uni-LSTM-1}]{\includegraphics[width=5.5cm]{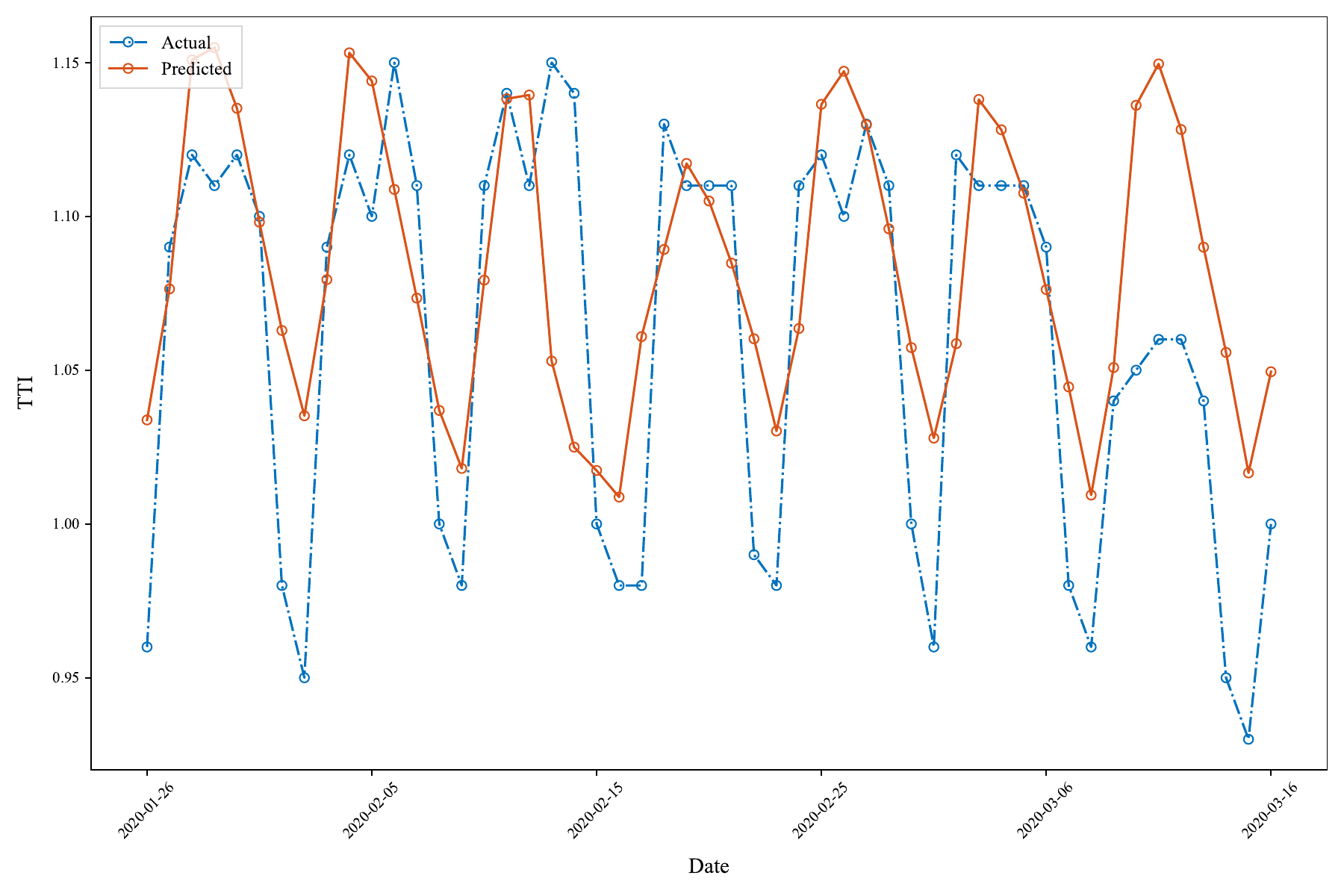}}
\subfloat[Bi-LSTM\label{subfig:Bi-LSTM-1}]{\includegraphics[width=5.5cm]{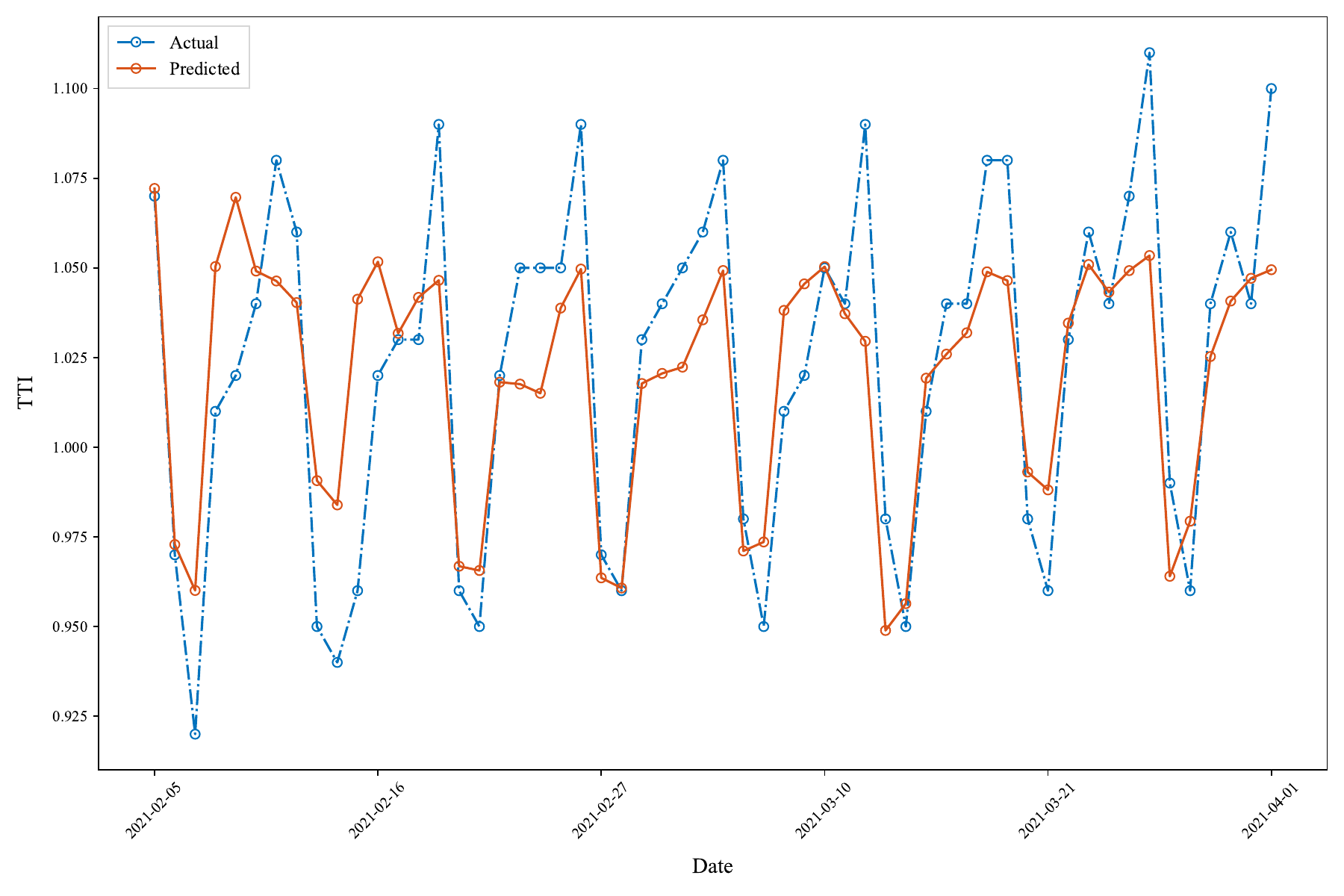}}
\caption{Comparison of Actual vs. Predicted TTI for Period 1}
\label{fig:vis-1}
\end{figure}

\begin{figure}
\centering
\subfloat[SVR\label{subfig:SVM-2}]{\includegraphics[width=5.5cm]{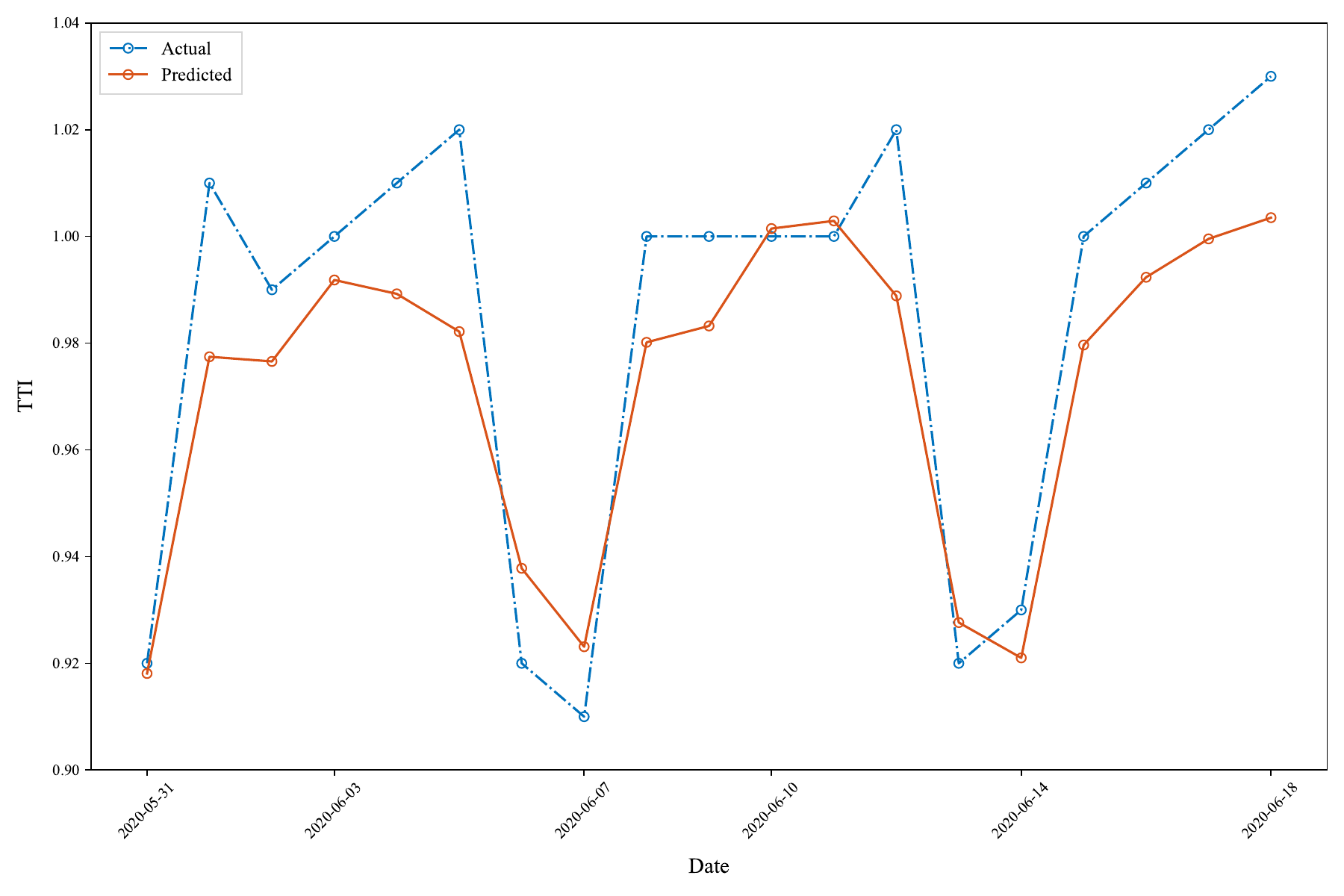}}
\subfloat[MLR\label{subfig:LR-2}]{\includegraphics[width=5.5cm]{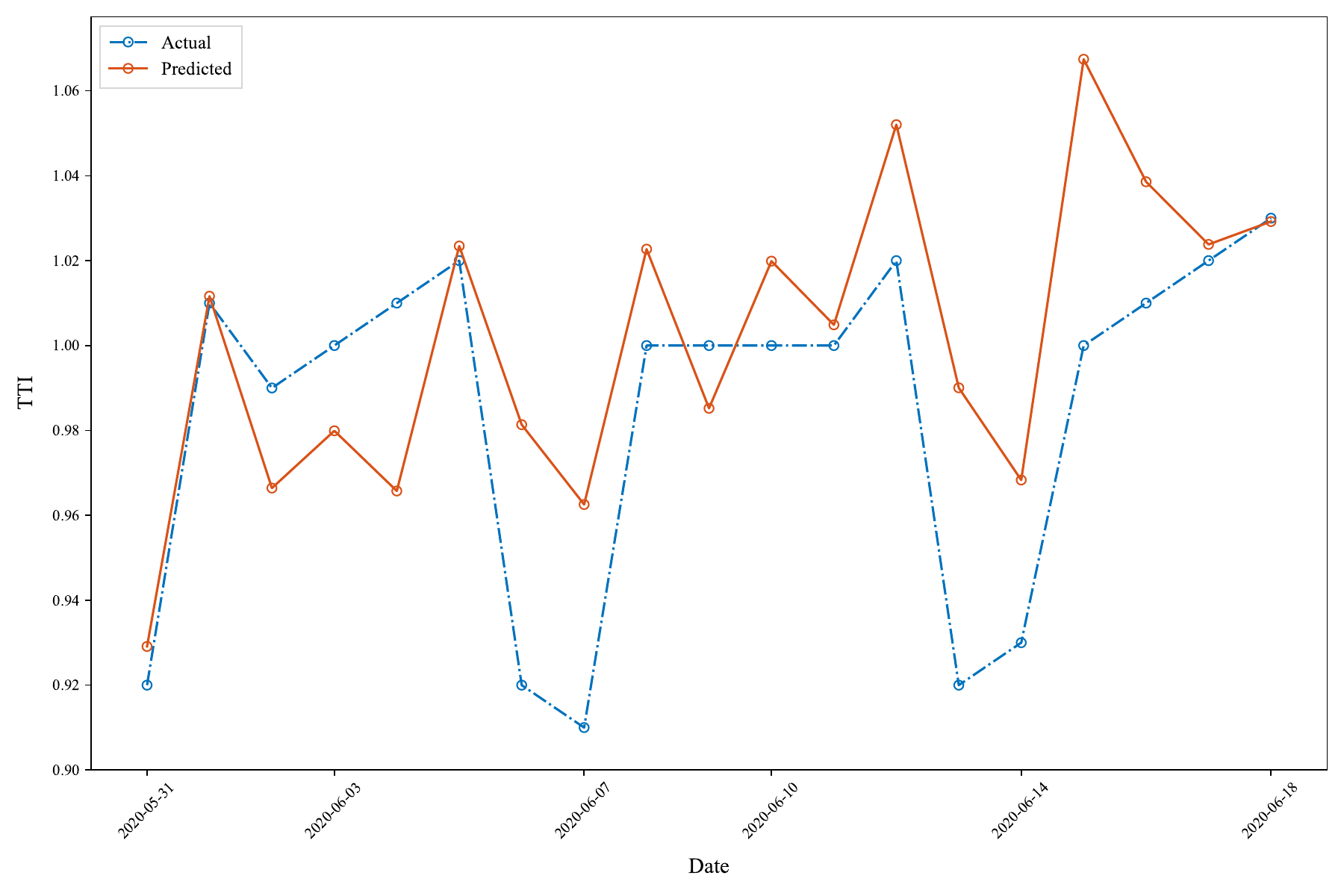}}
\subfloat[RNN\label{subfig:RNN-2}]{\includegraphics[width=5.5cm]{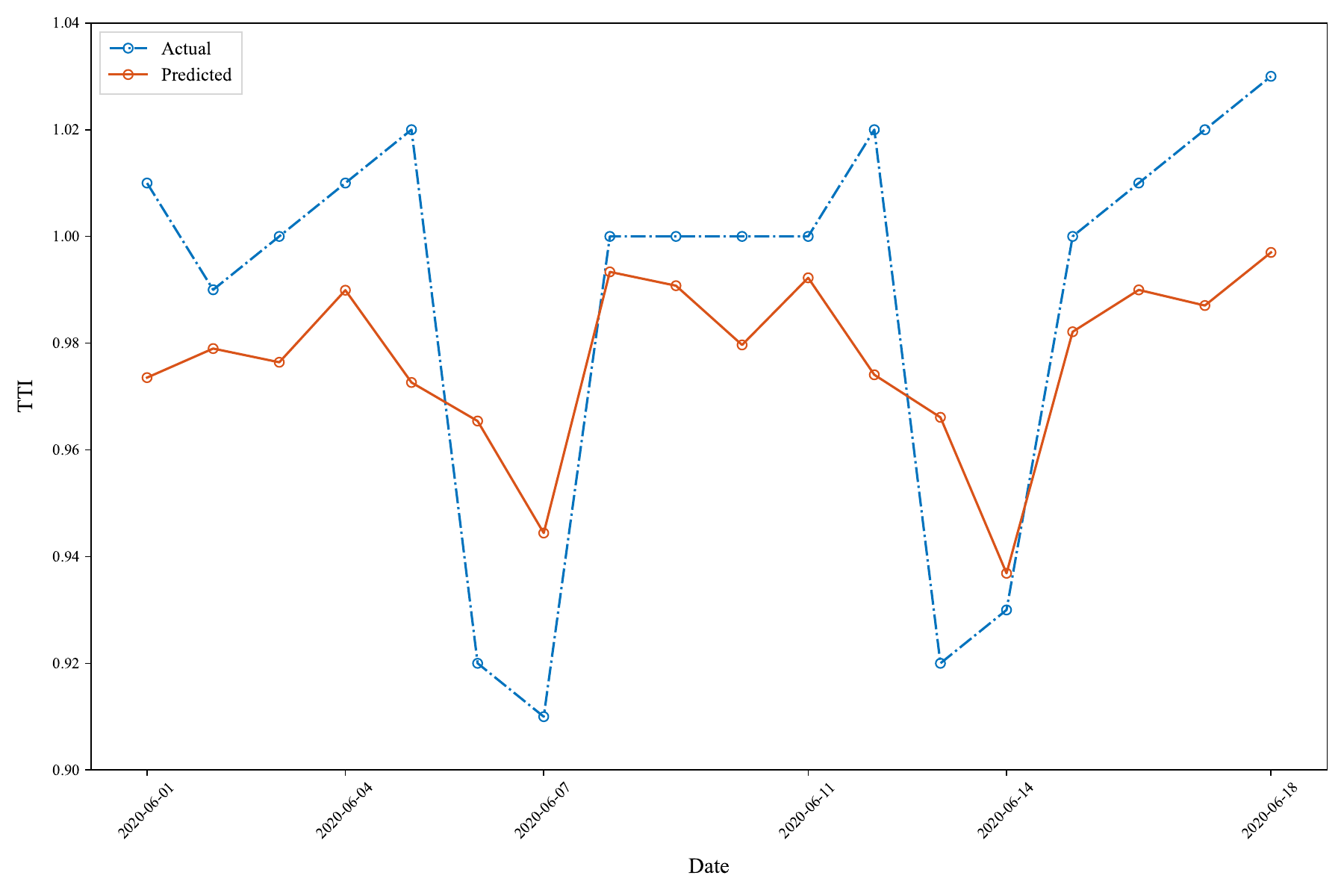}}\\
\subfloat[Uni-LSTM\label{subfig:Uni-LSTM-2}]{\includegraphics[width=5.5cm]{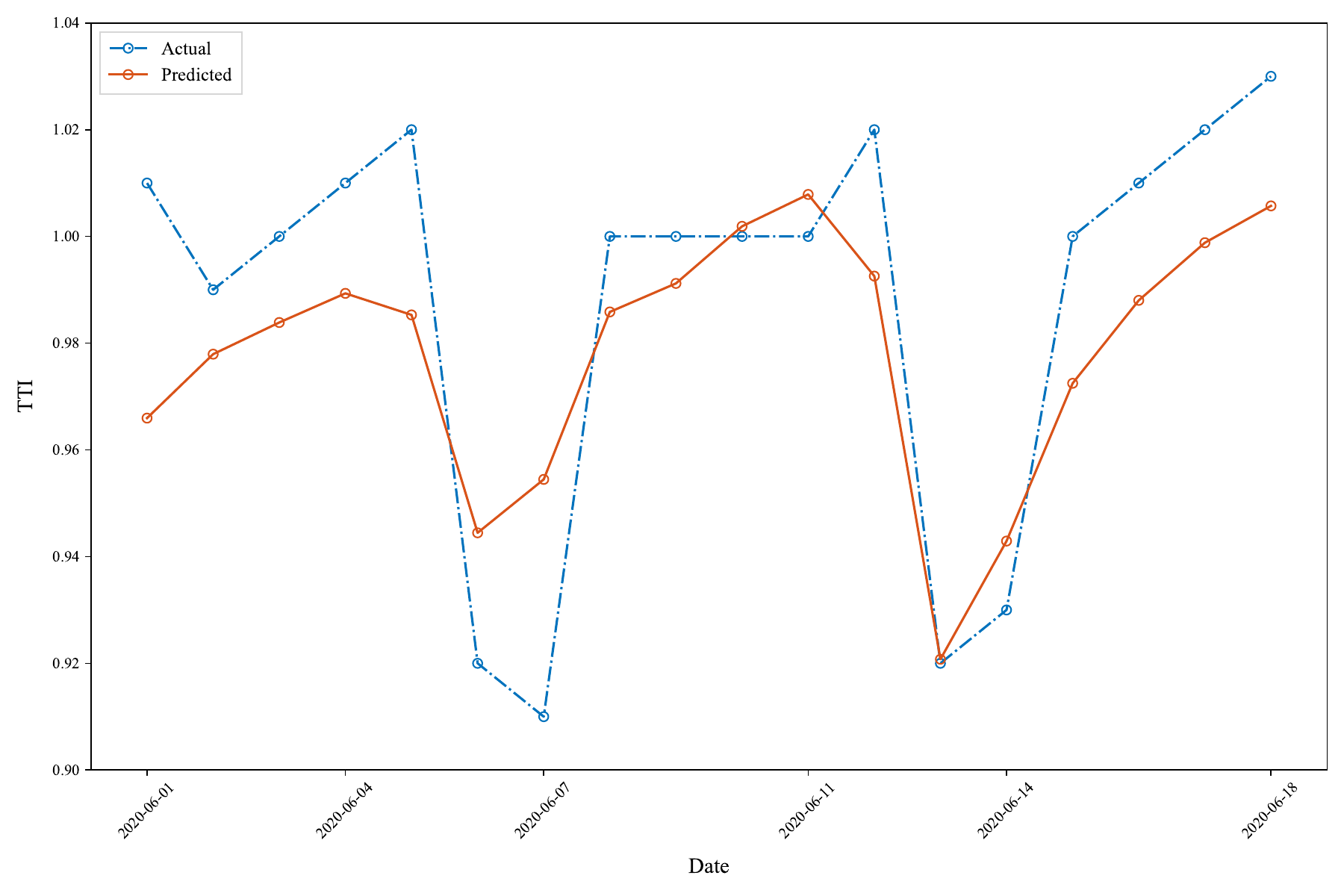}}
\subfloat[Bi-LSTM\label{subfig:Bi-LSTM-2}]{\includegraphics[width=5.5cm]{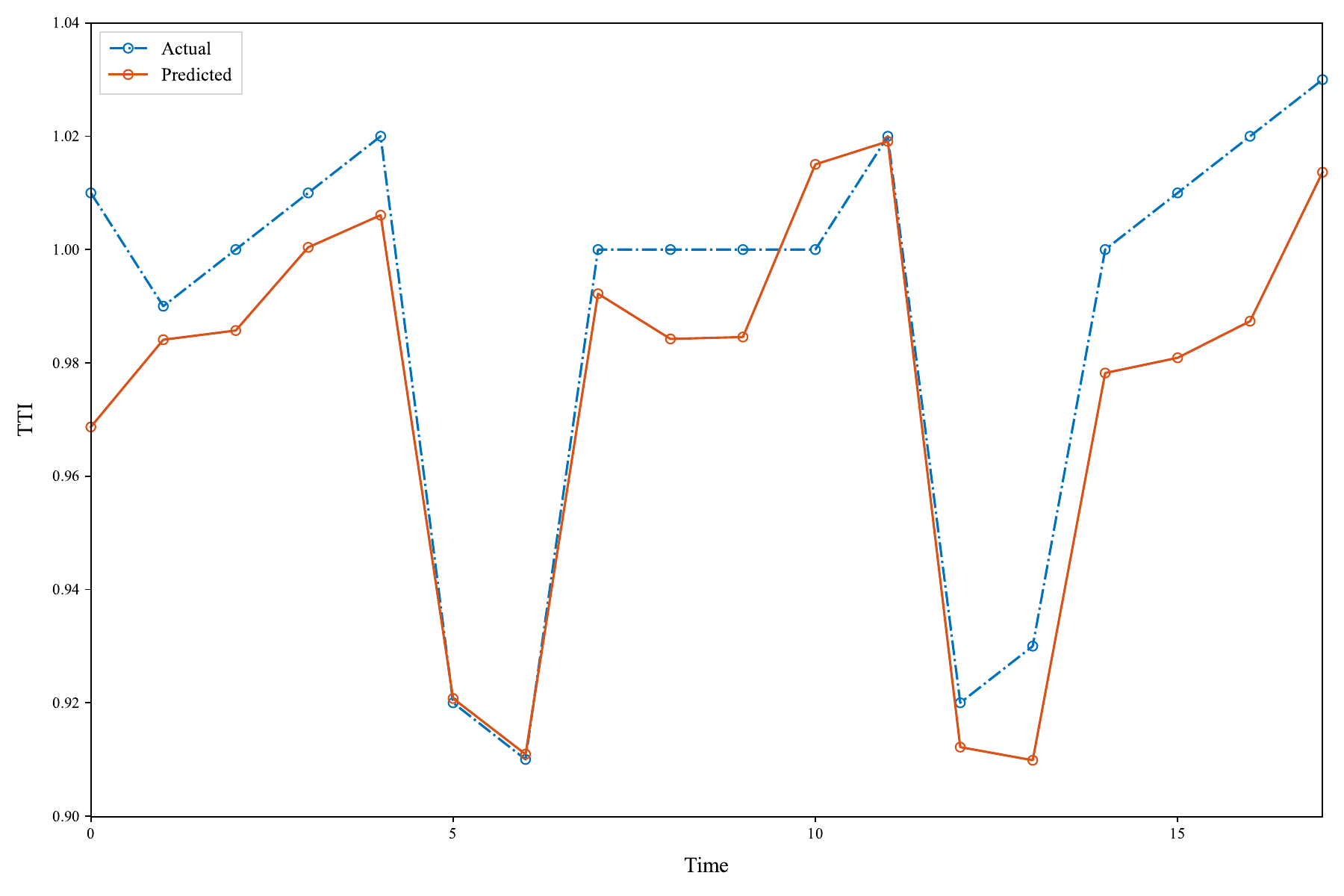}}
\caption{Comparison of Actual vs. Predicted TTI for Period 2}
\label{fig:vis-2}
\end{figure}

\begin{figure}
\centering
\subfloat[SVR\label{subfig:SVM-3}]{\includegraphics[width=5.5cm]{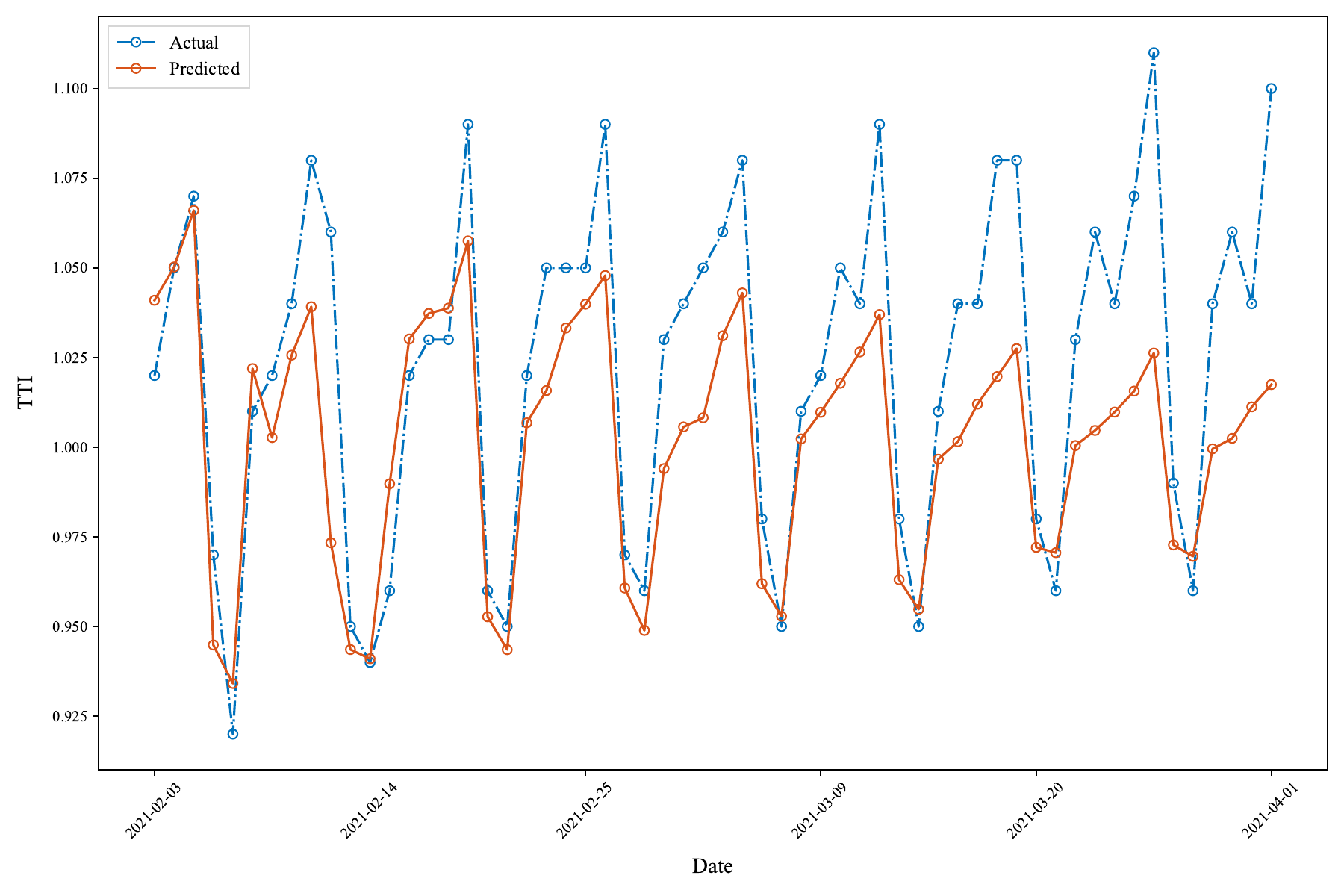}}
\subfloat[MLR\label{subfig:LR-3}]{\includegraphics[width=5.5cm]{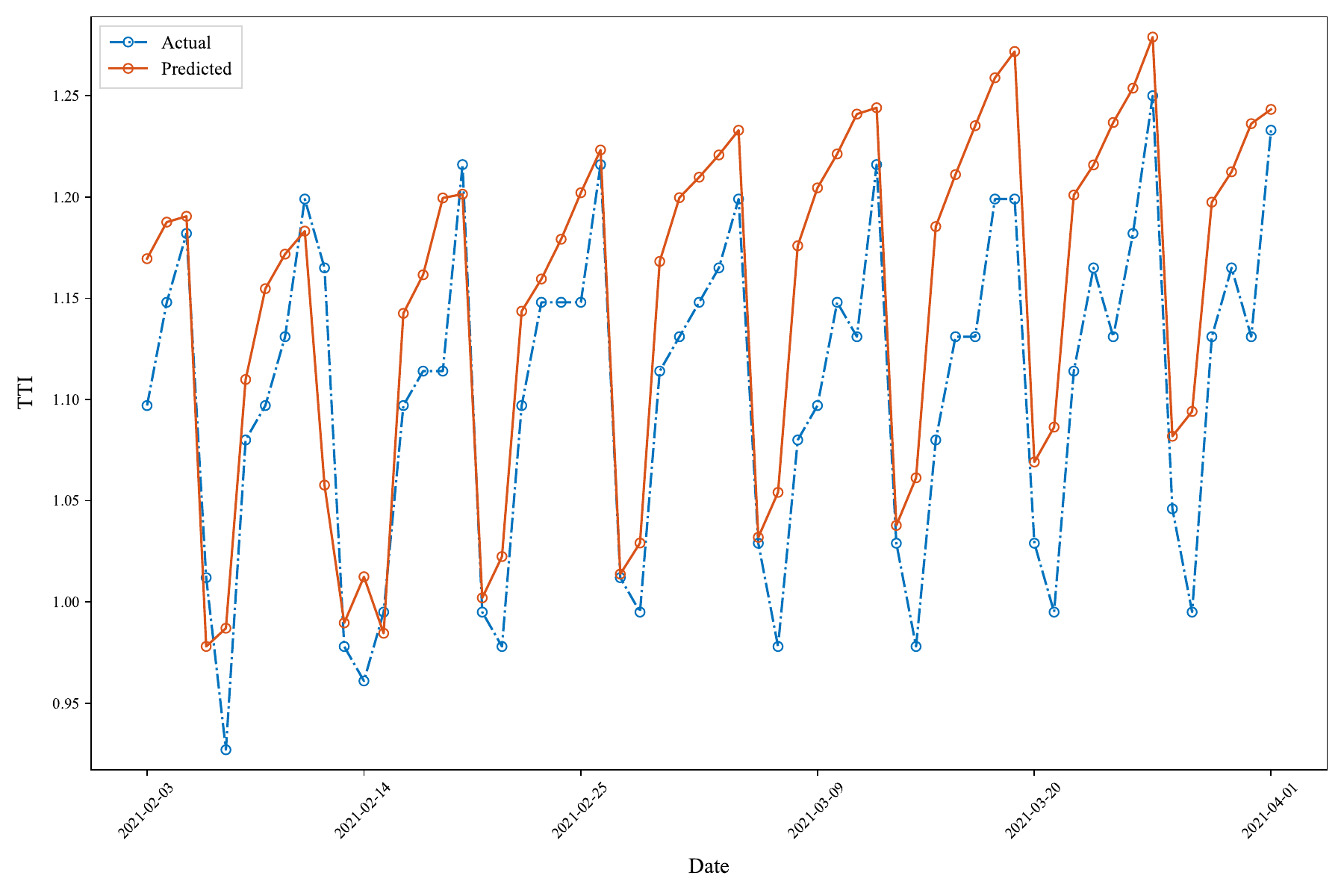}}
\subfloat[RNN\label{subfig:RNN-3}]{\includegraphics[width=5.5cm]{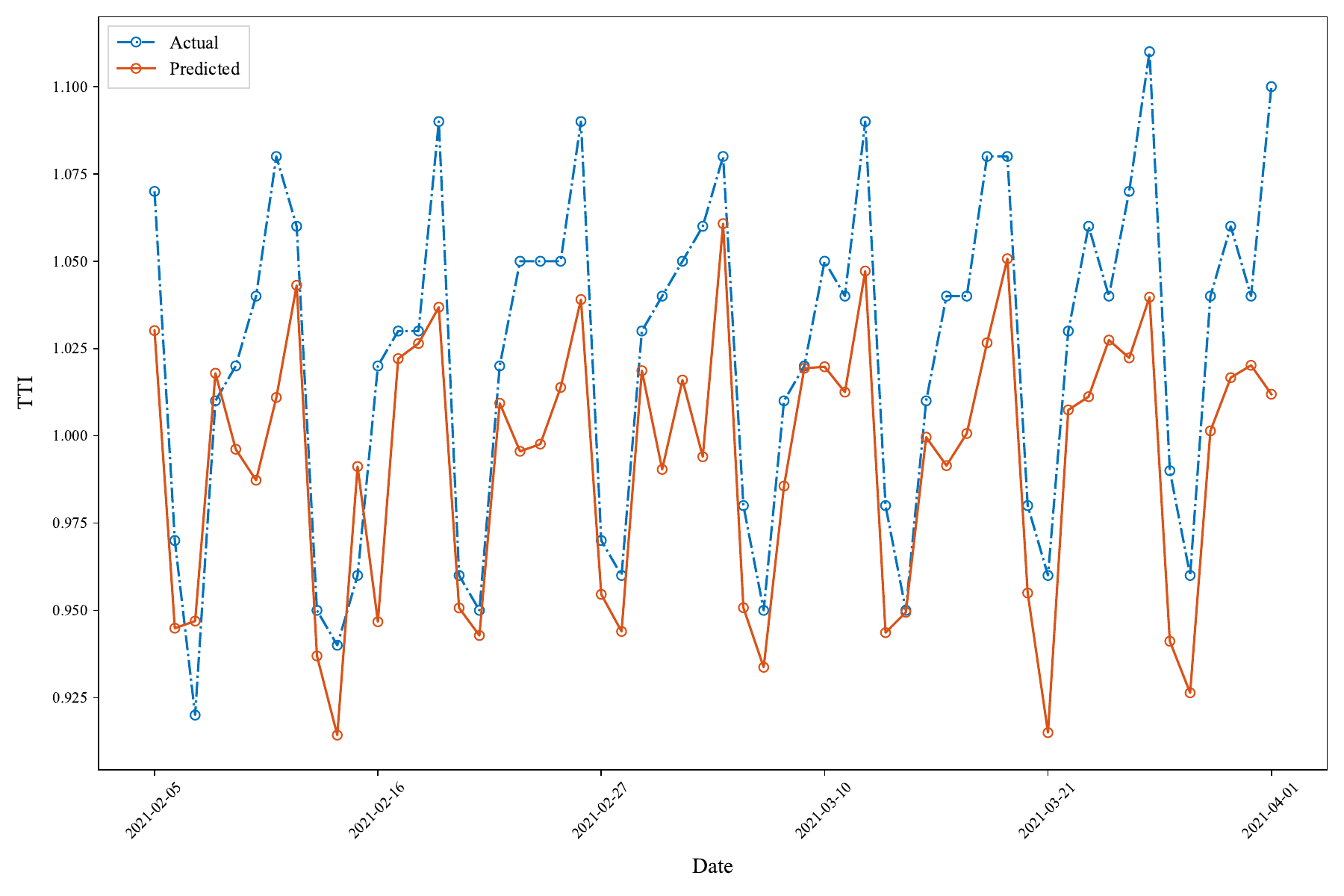}}\\
\subfloat[Uni-LSTM\label{subfig:Uni-LSTM-3}]{\includegraphics[width=5.5cm]{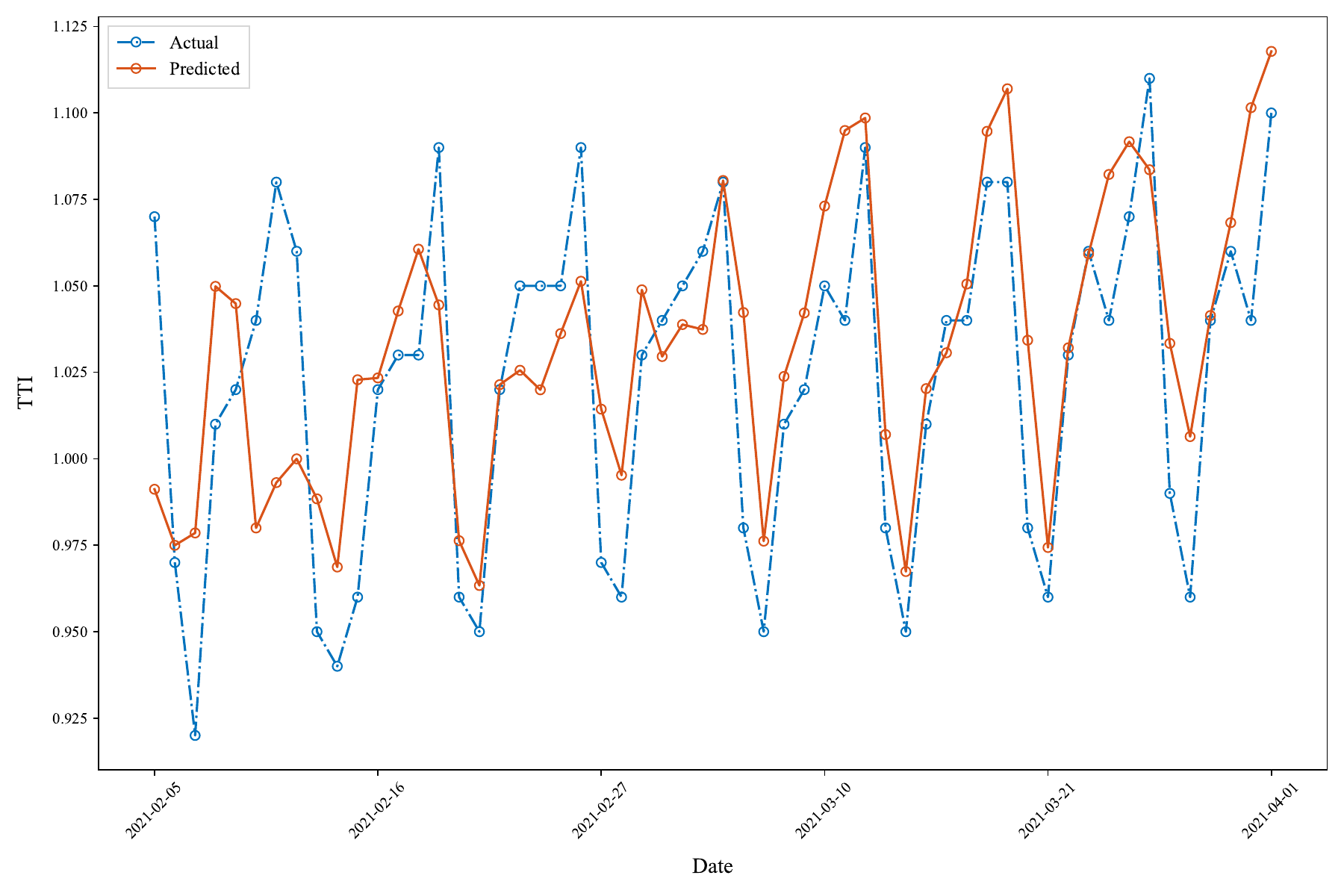}}
\subfloat[Bi-LSTM\label{subfig:Bi-LSTM-3}]{\includegraphics[width=5.5cm]{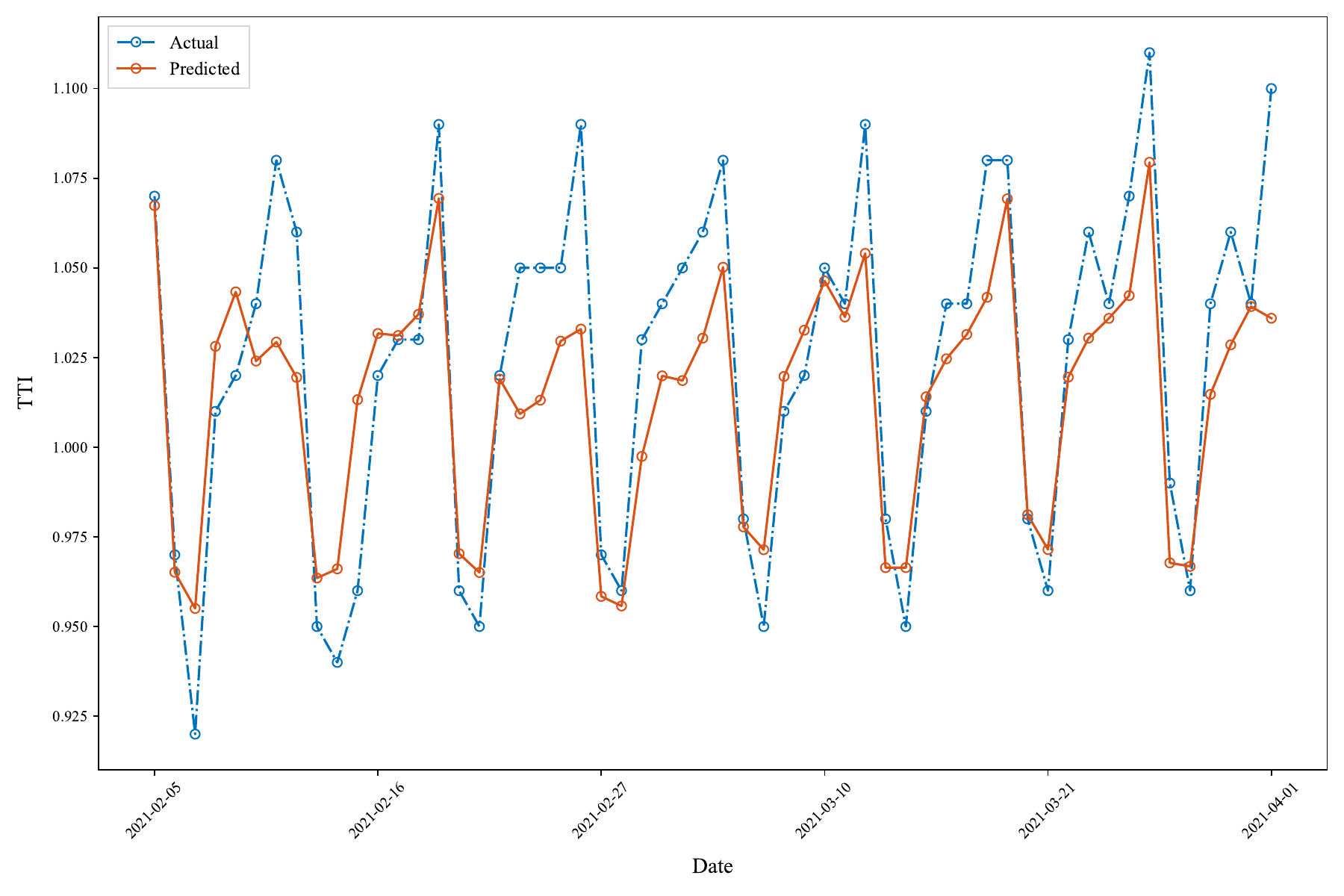}}
\caption{Comparison of Actual vs. Predicted TTI for Period 3}
\label{fig:vis-3}
\end{figure}

To further demonstrate the prediction details for different ML models over time. Figures \ref{fig:vis-1}, \ref{fig:vis-2}, and \ref{fig:vis-3} compare actual vs. predicted traffic values over time for periods 1, 2, and 3. {\color{blue}It can be readily seen that the Bi-LSTM model has the best predictive performance based on error criterion NRMSE, as the trend lines for the Bi-LSTM models are consistently closer to the actual values, which explains why the Bi-LSTM model has the lowest NRMSE in all periods.}

Besides, these three figures also allow us to compare the accuracy of the traffic prediction models developed for three distinct periods. Overall, the pre-lockdown period proves to be the most challenging to predict, followed by the post-lockdown period, with the lockdown period being the easiest to forecast. Remarkably, Period 2 emerges as the pinnacle of predictive accuracy among the tested models. During this period, stringent lockdown measures leads to a uniformity in traffic flow, with fewer vehicles on the road,  and the usual rush hour peaks is significantly flattened. Consequently, traffic dynamics are considerably more predictable because of the minimization of typical traffic influencers such as varying commute patterns and discretionary travel. Therefore, all models for this period achieve lower NRMSE scores, with the Bi-LSTM model showcasing the highest accuracy. The challenge escalates in Period 3, where the traffic landscape was influenced by a mix of gradual lifting of restrictions, the progress of vaccination campaigns, and a shift in public behavior. These factors introduce new variability into traffic patterns, making accurate predictions more demanding. Despite these complexities, the ability of Bi-LSTM to capture bidirectional temporal dependencies enables it to achieve the lowest NRMSE, highlighting its adaptability to changing conditions.

Lastly, during period 1, traffic patterns displayed significantly higher variability than the lockdown period. This increased fluctuation, unlike the pandemic's consistent impact, resulted from a confluence of factors including economic activity, weather, and seasonal trends. In contrast, Period 2, characterized by its shorter duration (94 records vs. 260 for Period 1, as detailed in Section 3.2), exhibited reduced sensitivity to weather and seasonal influences. This suggests that these non-COVID-related variables introduce considerable unpredictability, rendering Period 1 the most challenging for traffic prediction. The elevated NRMSE values observed across all models corroborate this difficulty.

\subsubsection{Benchmarking ML Methods against Seasonal Auto-Regressive Integrated Moving Average}
To further benchmark the effectiveness of the ML models, we compare their performance with a widely used traditional time series model, the Seasonal AutoRegressive Integrated Moving Average (SARIMA). SARIMA is an extension of the ARIMA model introduced by Box and Jenkins in the 1970s \citep{box2015time}, designed to handle time series data with seasonal patterns. It incorporates autoregressive, differencing, and moving average components, along with additional seasonal terms to capture repeating cycles in the data. The model assumes a linear relationship in the time series and typically performs well when the underlying patterns are stationary or seasonally stable. We choose SARIMA as a baseline because it remains one of the most widely used statistical forecasting models in transportation and time series analysis \citep{arunkumar2021forecasting}. A comparison of Table~\ref{Tab:SARIMA} and Figure~\ref{fig:per_total} shows that, although SARIMA performs slightly better than ML models in the pre-COVID period, it fails to outperform the Bi-LSTM and SVR models during the COVID period, as well as the Bi-LSTM model in the post-COVID period. This suggests that ML models are more flexible and better suited to capturing the complex dynamics of traffic congestion during and after large-scale disruptions such as a pandemic. While it is reasonable for SARIMA to perform well under stable, seasonal conditions—as in the pre-COVID period—its rigid structure limits its adaptability when the data exhibit irregular patterns or abrupt changes. In contrast, ML models, particularly Bi-LSTM, can effectively learn long-term dependencies and adapt to sudden shifts in behavior, providing more accurate predictions in such contexts.
\begin{table*}[!ht]
\caption{Prediction results for SARIMA across three periods}
\centering
\begin{tabular}{cccc}
\toprule
\makecell[c]{} & \makecell[c] {Period 1} & \makecell[c] {Period 2} & \makecell[c] {Period 3}\\ 
\midrule 
\makecell[c]{NRMSE} & \makecell[c] {0.0279} & \makecell[c] {0.0235} & \makecell[c] {0.0258}\\
\bottomrule
\end{tabular}
\label{Tab:SARIMA}
\end{table*}

Moreover, one of the key advantages of ML models lies not only in predictive accuracy but also in interpretability. By incorporating methods such as SHapley Additive exPlanations (SHAP) and  Integrated Gradients (IG) into these models, we are able to identify how external factors (\emph{e.g.}, seasonal, weather, and COVID-19 related factors)  take effect on driving traffic congestion, as illustrated in the following section. On the contrary, this explanatory capability is not available in traditional models like SARIMA.  To summarize, we recommend using ML models over SARIMA due to their superior adaptability to irregular and dynamic traffic patterns, as well as their added advantage of interpretability through techniques such as SHAP and Integrated Gradients.

\subsubsection{Prediction Performance under Different Traffic Conditions}
To further investigate the temporal and contextual sensitivity of the prediction models, we evaluate their performance under varying traffic conditions. Specifically, we compare model accuracy between weekday vs. weekend and $precipitation > 0$ vs. $= 0$ scenarios. This section presents the findings based on a representative model (\emph{i.g.,} MLR), and discusses potential implications for model deployment in real-world settings. We believe the prediction accuracy under other models will be similar, to avoid redundancy, we omit it here.

Tables~\ref{Tab:WEEK} and~\ref{Tab:PRECI} present the prediction accuracy of MLR model across three time periods under different temporal and weather conditions. The performance is evaluated using NRMSE. 
Specifically, Table~\ref{Tab:WEEK} compares the MLR model's performance between weekday and weekend conditions. Across all three periods, the model achieves consistently lower prediction errors during weekends. Notably, in Period~2, the weekend error is as low as 0.0084, substantially lower than the corresponding weekday error of 0.0201. These results suggest that the model captures traffic patterns more effectively on weekends. This may be attributed to more regular and consistent travel behavior (e.g., leisure and family activities), resulting in reduced variability. Conversely, weekday travel is typically more complex due to commuting peaks, varying schedules, and mixed-purpose trips, which pose greater challenges for a linear predictive model.

\begin{table*}[!ht]
\caption{LR prediction accuracy by weekday/weekend across three periods}
\centering
\begin{tabular}{cccc}
\toprule
\makecell[c]{} & \makecell[c] {Period 1} & \makecell[c] {Period 2} & \makecell[c] {Period 3}\\ 
\midrule 
\makecell[c]{Weekend} & \makecell[c] {0.0161} & \makecell[c] {0.0084} & \makecell[c] {0.0253}\\
\midrule
\makecell[c]{Weekday} & \makecell[c] {0.0387} & \makecell[c] {0.0201} & \makecell[c] {0.0369}\\
\bottomrule
\end{tabular}
\label{Tab:WEEK}
\end{table*}
Table~\ref{Tab:PRECI} evaluates the model's accuracy under varying precipitation conditions. In Periods~1 and~2, the model performs better when there is no precipitation, indicating that rain introduces more variability and unpredictability into travel behavior, thereby reducing prediction accuracy. Interestingly, in Period~3, the trend reverses: the prediction accuracy under precipitation (0.0177) is lower than that under dry conditions (0.0376). One possible explanation is that travel demand during this period became more regulated or reduced---potentially due to post-pandemic behavioral changes such as increased remote working, reduced discretionary travel, and more cautious travel behavior---resulting in more predictable patterns even under adverse weather conditions.
 
\begin{table*}[!ht]
\caption{LR prediction accuracy by precipitation level across three periods}
\centering
\begin{tabular}{cccc}
\toprule
\makecell[c]{} & \makecell[c] {Period 1} & \makecell[c] {Period 2} & \makecell[c] {Period 3}\\ 
\midrule 
\makecell[c]{$precipitation >0$} & \makecell[c] {0.0433} & \makecell[c] {0.0468} & \makecell[c] {0.0177}\\
\midrule
\makecell[c]{$precipitation = 0$} & \makecell[c] {0.0333} & \makecell[c] {0.0398} & \makecell[c] {0.0376}\\
\bottomrule
\end{tabular}
\label{Tab:PRECI}
\end{table*}
Overall, the results highlight that the MLR model's performance is sensitive to both temporal and environmental heterogeneity. The model performs best under stable and regular traffic conditions (e.g., weekends and dry weather), and less effectively when behavioral variability is higher (e.g., weekdays and rain). 

\subsection{Model Interpretability}
In this section, we explain how the input features impact the TTI for the Bi-LSTM model via IG and the SVR via SHAP values.

\subsubsection{Integrated Gradients}
\begin{figure}
\centering
\subfloat[a zero baseline\label{subfig:IG-1}]{\includegraphics[width=7.2cm]{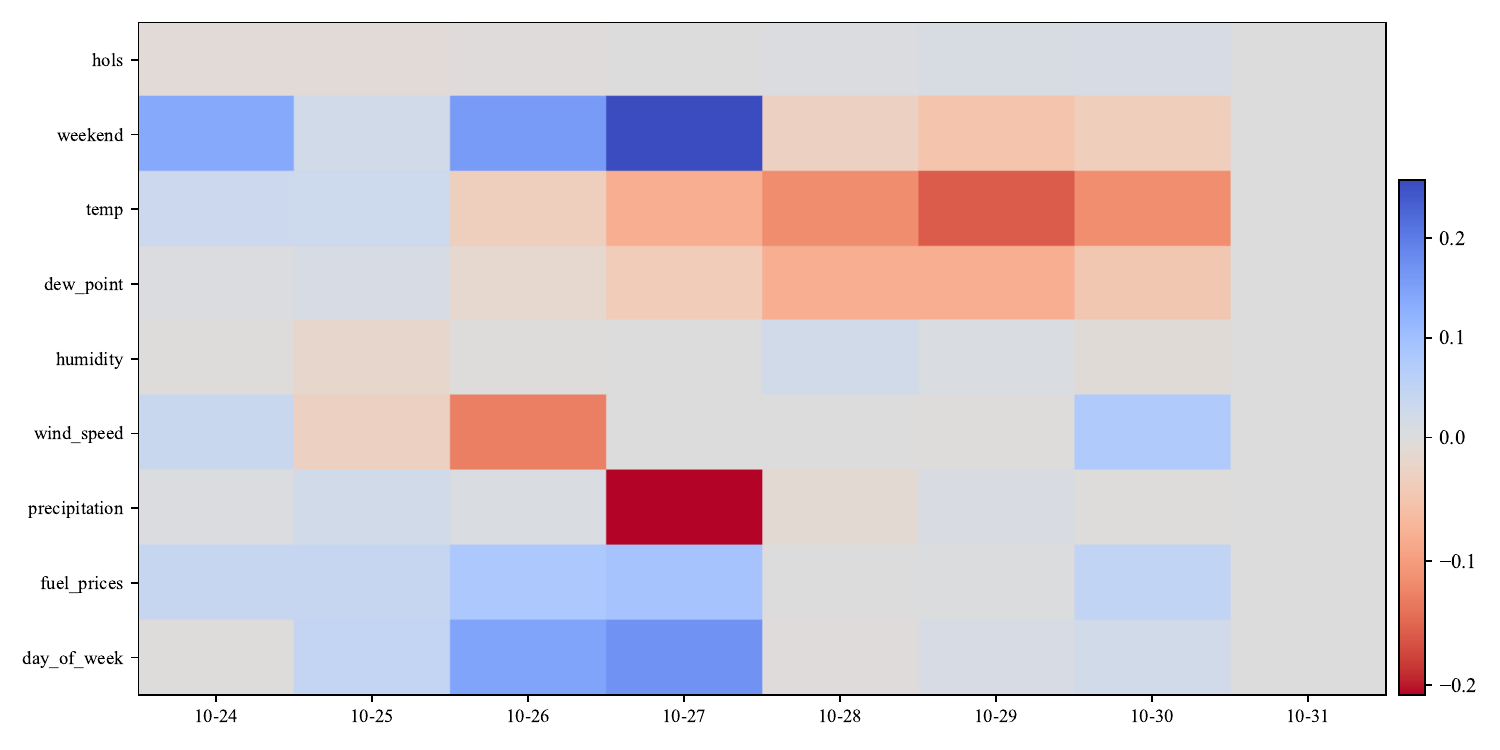}}
\subfloat[a mean baseline \label{subfig:IG-2}]{\includegraphics[width=7.2cm]{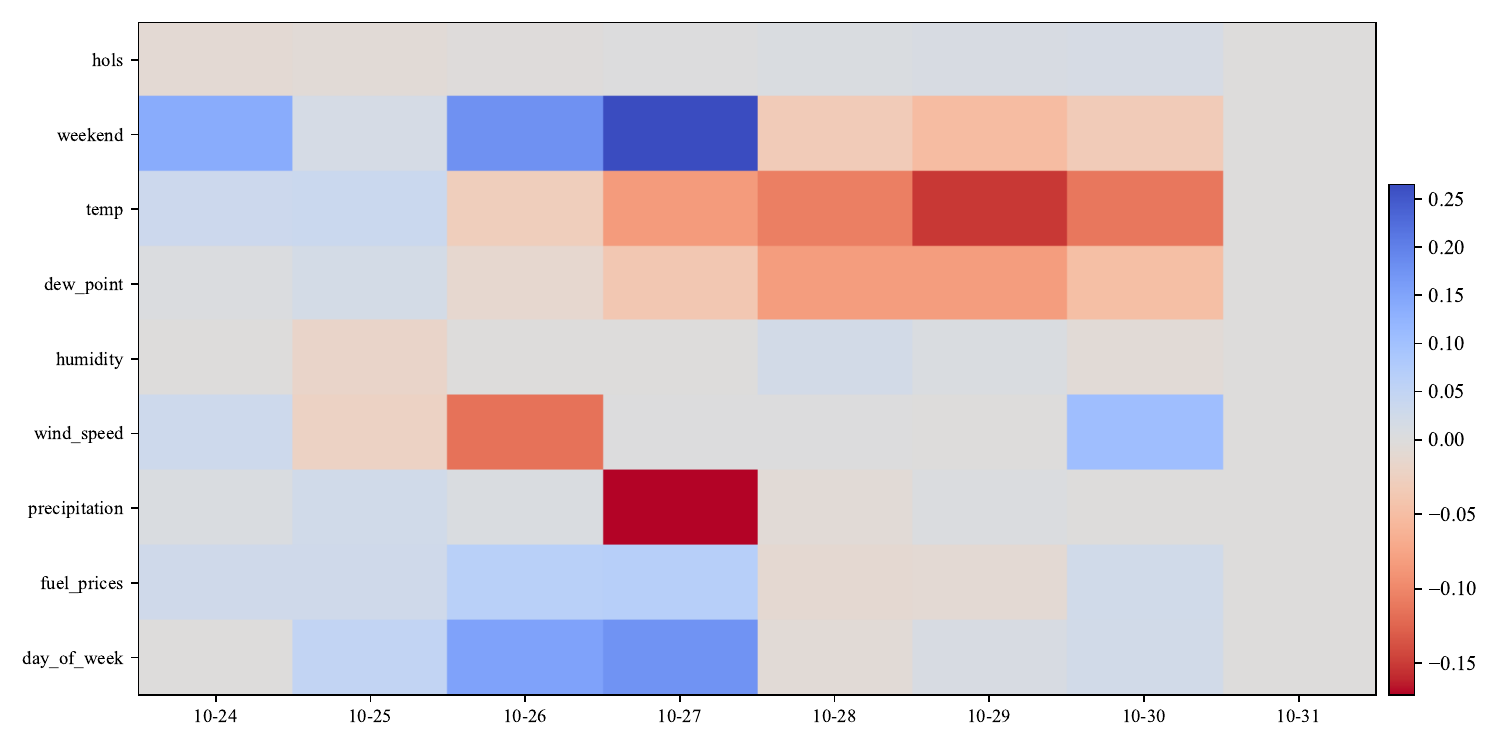}}\\
\subfloat[a past-week baseline\label{subfig:IG-3}]{\includegraphics[width=7.2cm]{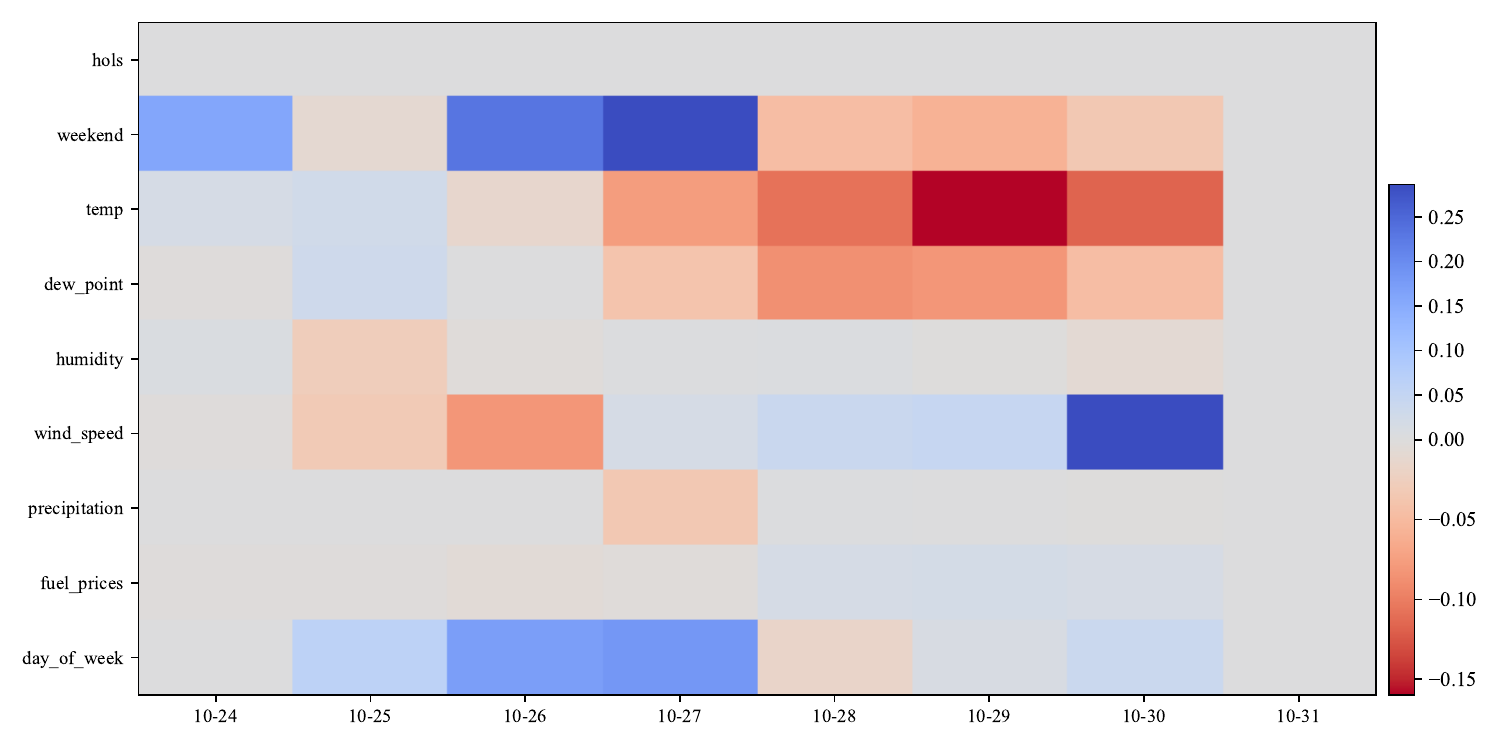}}
\caption{IG attribution maps of the Bi-LSTM model for pre-lockdown period (2019-10-31) }
\label{fig:IG-map-ZERO}
\end{figure}

\begin{figure}
\centering
\subfloat[a zero baseline\label{subfig:IG-1}]{\includegraphics[width=7.2cm]{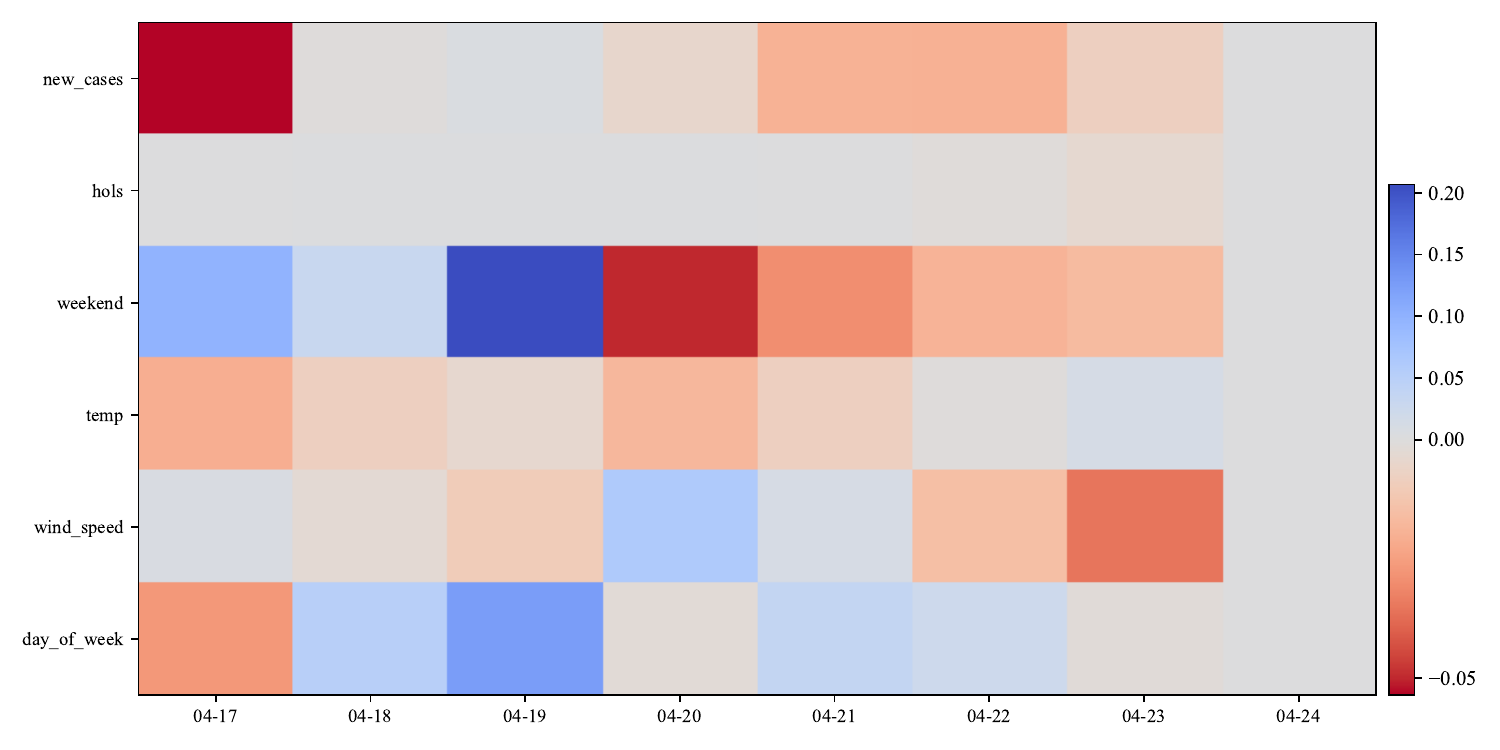}}
\subfloat[a mean baseline\label{subfig:IG-2}]{\includegraphics[width=7.2cm]{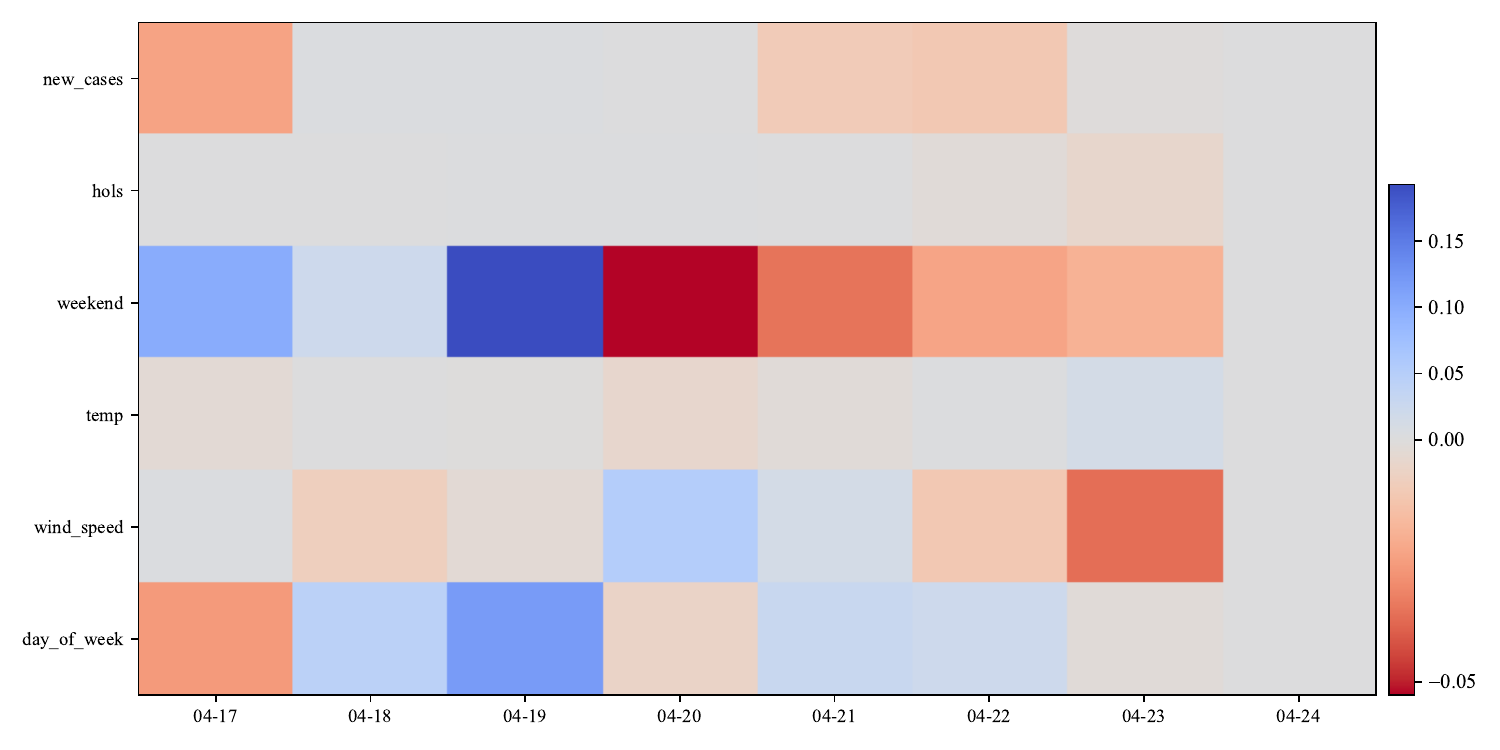}}\\
\subfloat[a past-week baseline\label{subfig:IG-3}]{\includegraphics[width=7.2cm]{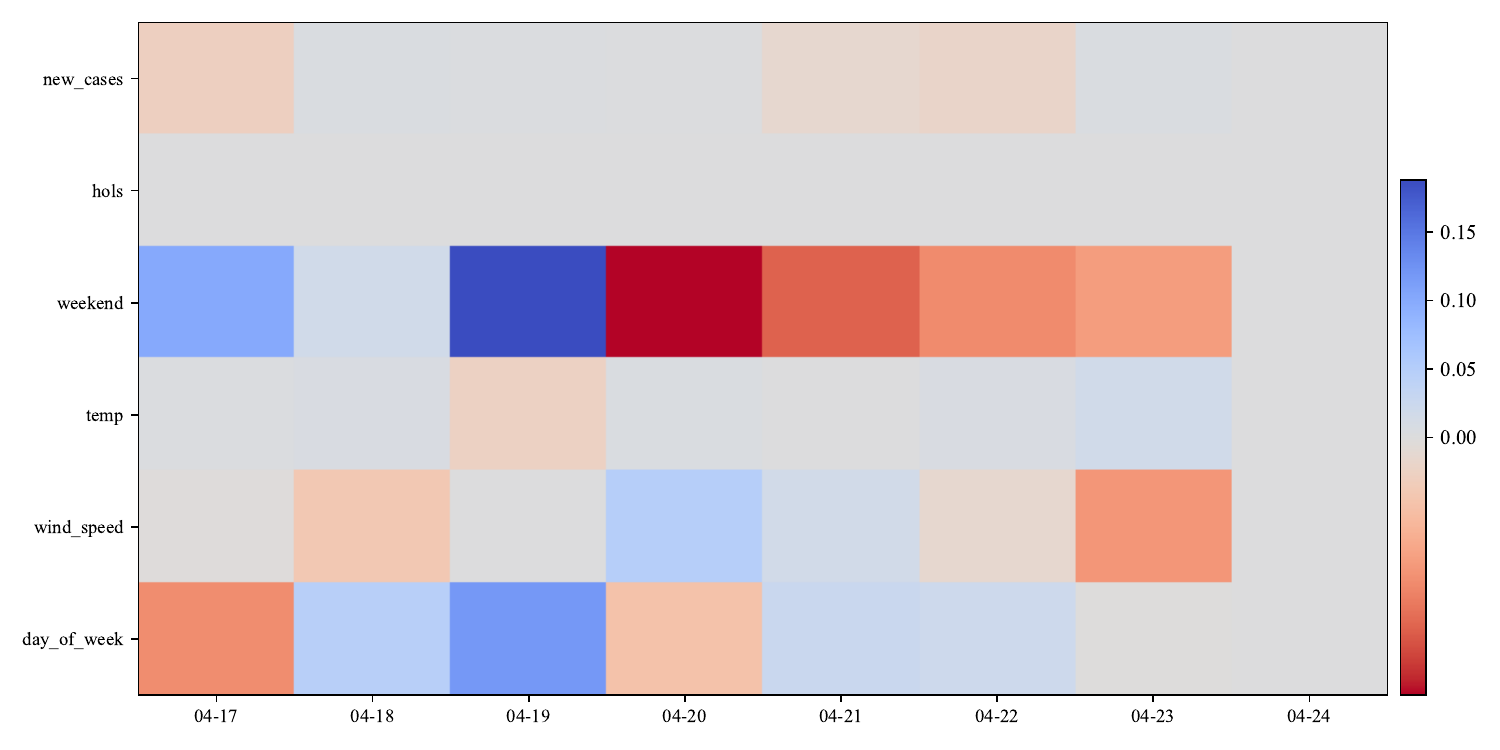}}
\caption{IG attribution maps of the Bi-LSTM model for the lockdown period (2020-04-24)}
\label{fig:IG-map-MEAN}
\end{figure}

\begin{figure}
\centering
\subfloat[a zero baseline\label{subfig:IG-1}]{\includegraphics[width=7.2cm]{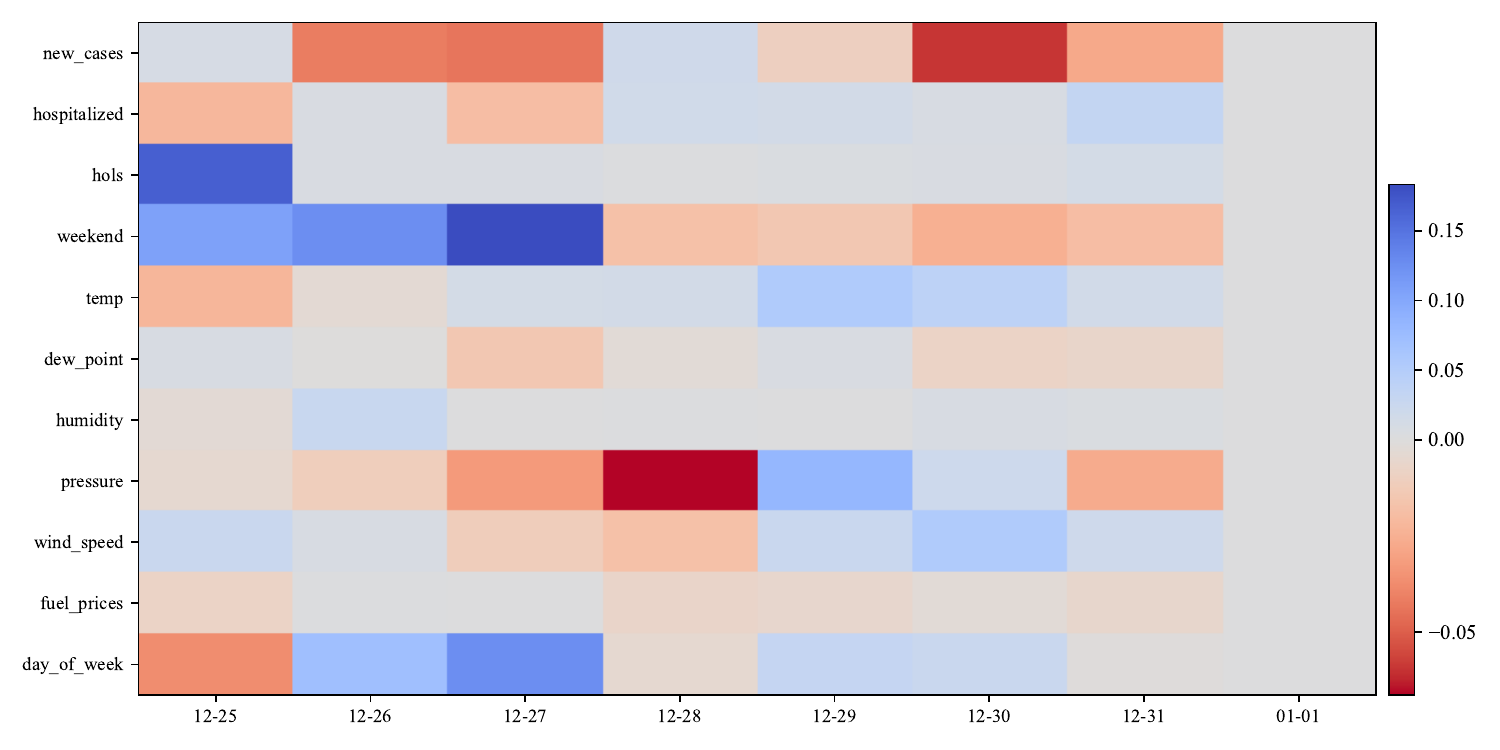}}
\subfloat[a mean baseline\label{subfig:IG-2}]{\includegraphics[width=7.2cm]{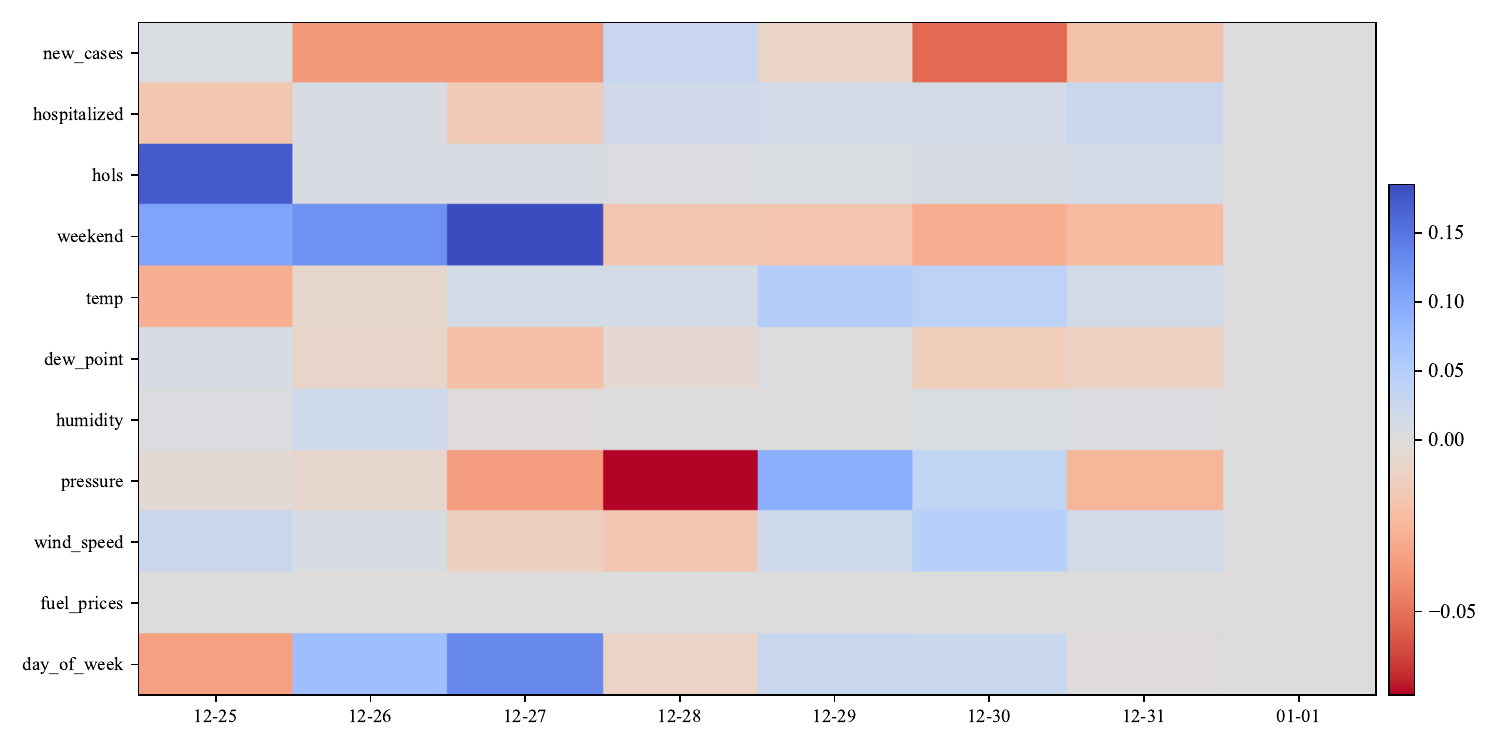}}\\
\subfloat[a past-week baseline\label{subfig:IG-3}]{\includegraphics[width=7.2cm]{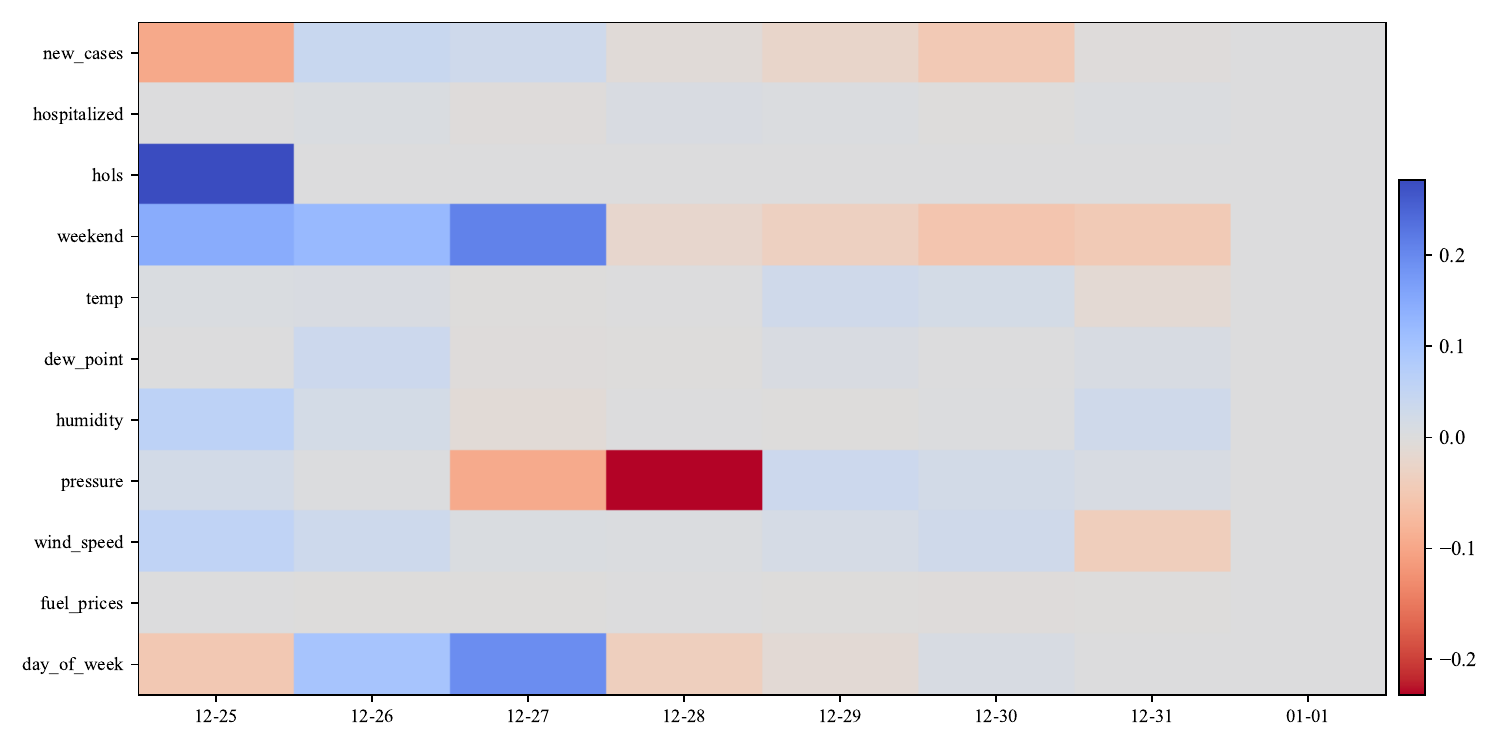}}
\caption{IG attribution maps of the Bi-LSTM model for the post-lockdown period (2021-01-01)}
\label{fig:IG-map-PASTWEEK}
\end{figure}

The interpretability of machine learning models is essential for understanding the decision-making process, especially in domains such as urban traffic management. In this study, we consider Integrated Gradients (IG), which is a technique designed to offer insights into how different input features impact a model's predictions. It works by attributing the prediction output to the input features based on their gradients, which are computed by creating a path from a baseline (where the feature is absent) to the actual input. This method can be applied to any differentiable model that includes various types of neural networks. We apply it to the Bi-LSTM model, as it outperforms all other differentiable models, including RNN and Uni-LSTMs. We adopt a 7-day lookback window for IG analysis, which allows for more interpretable IG attributions, as the model can assign importance scores to specific days within a complete weekly context. Note that IG computes the attributions by integrating gradients along a straight path from a baseline input to the actual input. The baseline represents a ``neutral'' or ``uninformative'' state - essentially what the model would predict in the absence of meaningful input. In this study, we consider three different baselines to test for attribution stability. A zero baseline assumes no feature activation, useful when input features are normalized and zero represents the absence of information. A mean baseline represents the average input, reflecting a typical or expected scenario. A past-week baseline captures short-term temporal trends and offers a dynamic, context-aware reference.
Figures \ref{fig:IG-map-ZERO}-\ref{fig:IG-map-PASTWEEK} display IG attribution heatmaps for pre-lockdown, lockdown, and post-lockdown periods, each shown under three baselines (zero, mean, past-week). The y-axis in each subfigure represents the variable names, while the x-axis represents the dates within the lookback window for each sample, with movement to the left indicating earlier points in time. Red (negative values) signifies a negative correlation between input from a specific lookback day and the target day, whereas blue indicates a positive correlation.

It is evident that features play dynamic roles across all periods. A common pattern observed in the attribution maps is that the middle of the lookback window consistently exhibits stronger attribution intensities, with influence gradually decaying toward both ends. This central concentration likely reflects the bidirectional memory mechanism of the Bi-LSTM model, which emphasizes balanced temporal dependencies. In addition, several features show shifting correlations with TTI, alternating between positive and negative values. For example, in Period 1, recent weekend indicators are negatively correlated with TTI, while older observations occasionally exhibit positive effects. Such variations highlight the complex and temporal nature of contributing factors, indicating that traffic congestion is shaped by a multifaceted interplay rather than a single dominant influence.

When comparing attribution maps under different baselines, we find a high degree of consistency across all three periods. Firstly, within each period, the same dominant features consistently emerge across the zero, mean, and past-week baselines. Secondly, the direction of each feature's influence on TTI (\emph{i.e.}, whether positive or negative) remains unchanged regardless of the baseline. Although the magnitude of attributions may vary with the choice of baseline, as each baseline sets a different reference point for gradient integration, this variation is a well-recognized and inherent characteristic of the Integrated Gradients method. 
Notably, attributions derived using the zero baseline tend to exhibit the strongest intensity, followed by the mean baseline, with the past-week baseline typically yielding the weakest attributions. This pattern arises because the zero baseline is more distant from the actual input values in the feature space, leading to larger gradients during integration. In contrast, the mean and past-week baselines are closer to the input distribution, resulting in more moderate gradients and thus less intense attributions.
Overall, the stability of dominant features and directional effects across baselines reinforces the robustness and interpretability of the IG-based explanation for our Bi-LSTM model.

When comparing IG results across the three periods, distinct patterns emerge that reflect evolving traffic conditions influenced by various external factors. In Period 1, seasonality features play a significant role-particularly the weekend indicator, which exhibits dynamic and shifting attribution patterns. Specifically, the weekend values show a positive influence on TTI in the earlier part of the lookback window (prior to October 27, 2019), but this influence transitions to negative as the prediction date approaches. This shift may reflect typical pre-pandemic travel behavior: earlier in the week, upcoming weekends might be associated with increased pre-weekend traffic buildup or event-driven travel, whereas traffic tends to subside on the actual weekend days themselves, resulting in lower congestion levels. This aligns with common weekly rhythms in traffic flow observed in our earlier trend analysis.
The model also responds notably to temperature, with this feature showing a consistently negative attribution across the entire lookback window. This suggests that higher temperatures during this period were associated with reduced traffic levels-possibly due to people avoiding non-essential travel during hot weather or shifting toward more comfortable, indoor activities.

In Period 2, with the onset of COVID-19 and the implementation of lockdown measures, the model's focus shifts. The emergence of new COVID-19 case counts is a dominant feature, consistently showing a strong negative attribution at all baselines. This negative relationship suggests that an increase in reported COVID-19 cases corresponds with a decrease in traffic levels, likely due to heightened public concern, voluntary avoidance behavior, or compliance with mobility restrictions. The prominence and directional stability of this feature across all baselines highlight its critical role during the lockdown phase and underscore the model's ability to capture the societal response to emerging health threats. Besides,  the weekend feature continues to play a prominent role, exhibiting a similar temporal pattern as observed in Period 1.

Period 3 shows a clear shift in feature dominance, with pandemic-related variables, such as new cases and hospitalizations, gaining more importance compared to traditional predictors. This suggests that during times of significant societal impact, such as the post-pandemic period, conventional factors are overshadowed by those directly tied to the event. A closer examination of Period 3 reveals that both the new COVID-19 cases and hospitalization variables exhibit predominantly negative attributions throughout the lookback window. This consistent negative contribution across time suggests that higher numbers of reported cases and hospitalizations are strongly associated with reduced traffic volumes. This reflects a more sustained behavioral response-such as continued remote work, heightened health precautions, or stricter public mobility guidelines-even in the post-lockdown phase.  Specifically, the new cases variable shows its strongest negative influence closer to the prediction day, suggesting that recent spikes in case numbers trigger a short-term behavioral response-such as increased caution or avoidance of public spaces-leading to a drop in traffic volume. This may reflect public sensitivity to up-to-date case reports or immediate policy changes, such as advisories or reimposed restrictions. In contrast, the hospitalization variable displays strong negative attribution at the earliest points in the lookback window, with its influence gradually fading as the prediction date approaches. This indicates a more long-term dampening effect on mobility,  arising from the sustained public perception of severity and continued institutional responses to high hospitalization rates. 
Additionally, the pressure variable emerges as a notable weather-related contributor in this period. It displays a clearly negative attribution, particularly in the recent days of the lookback window. This may reflect that lower pressure systems--often associated with poor weather--lead to decreased travel activity.

Overall, the application of IG offers valuable insights into the decision-making processes of our Bi-LSTM model for traffic prediction in the three distinct periods. By attributing importance to input features, IG clarifies the dynamic-and at times delayed-effects of variables such as new COVID-19 cases, new hospitalizations, weather conditions, and seasonal factors on predicted traffic volumes.

\subsubsection{SHapley Additive exPlanations}
\begin{figure}
    \centering
    \subfloat[\label{subfig:shap-be}]{%
        \includegraphics[width=7.2cm]{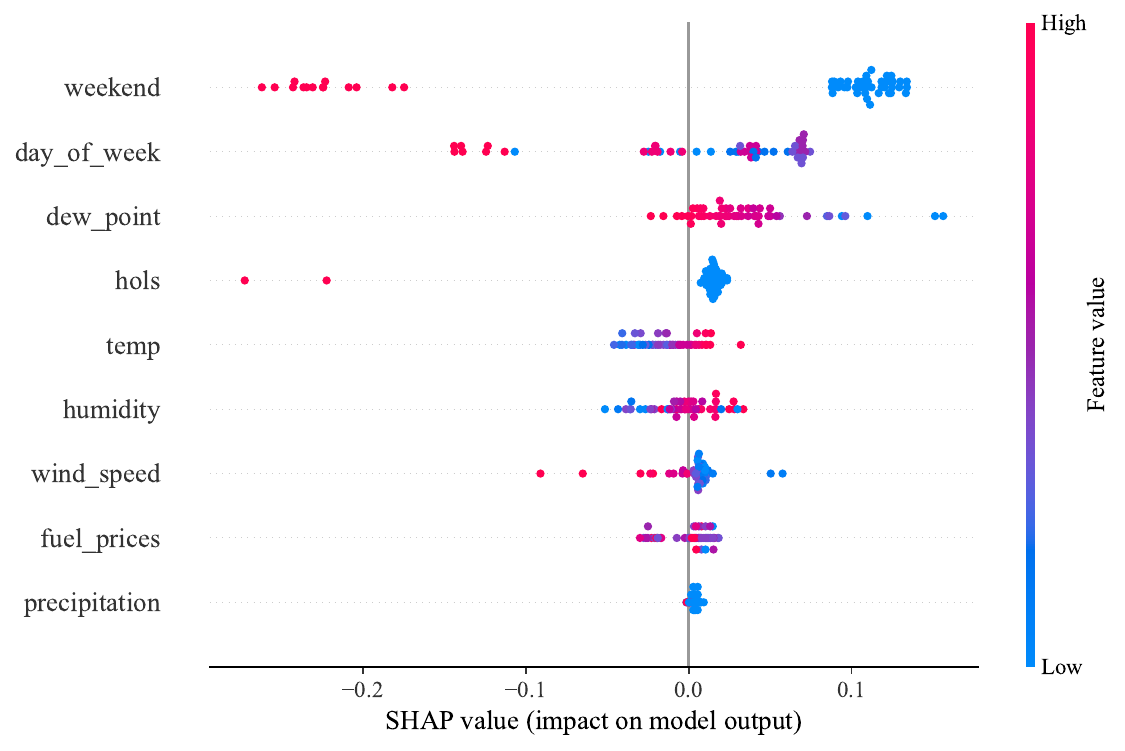}
    }
    \subfloat[\label{subfig:shap-af}]{%
        \includegraphics[width=6.2cm]{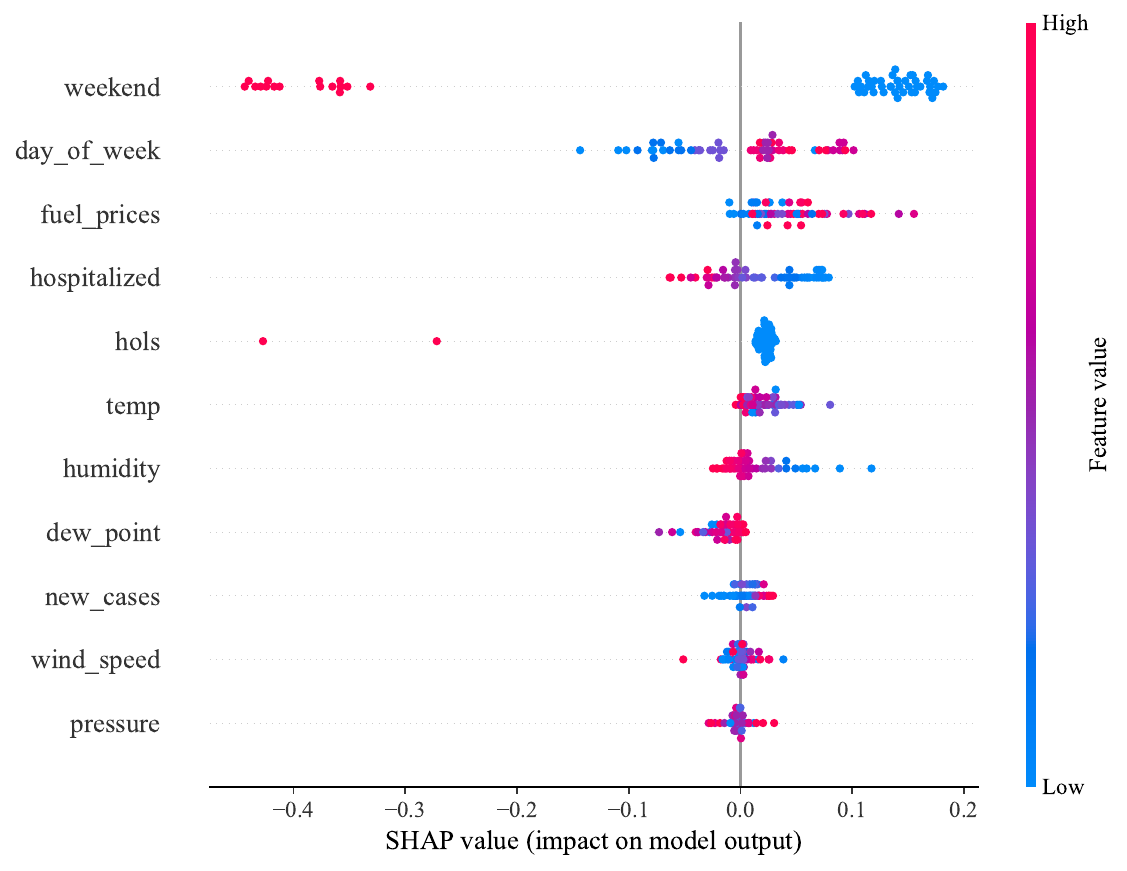}
    }\\
    \subfloat[\label{subfig:shap_gl}]{%
        \includegraphics[width=7.5cm]{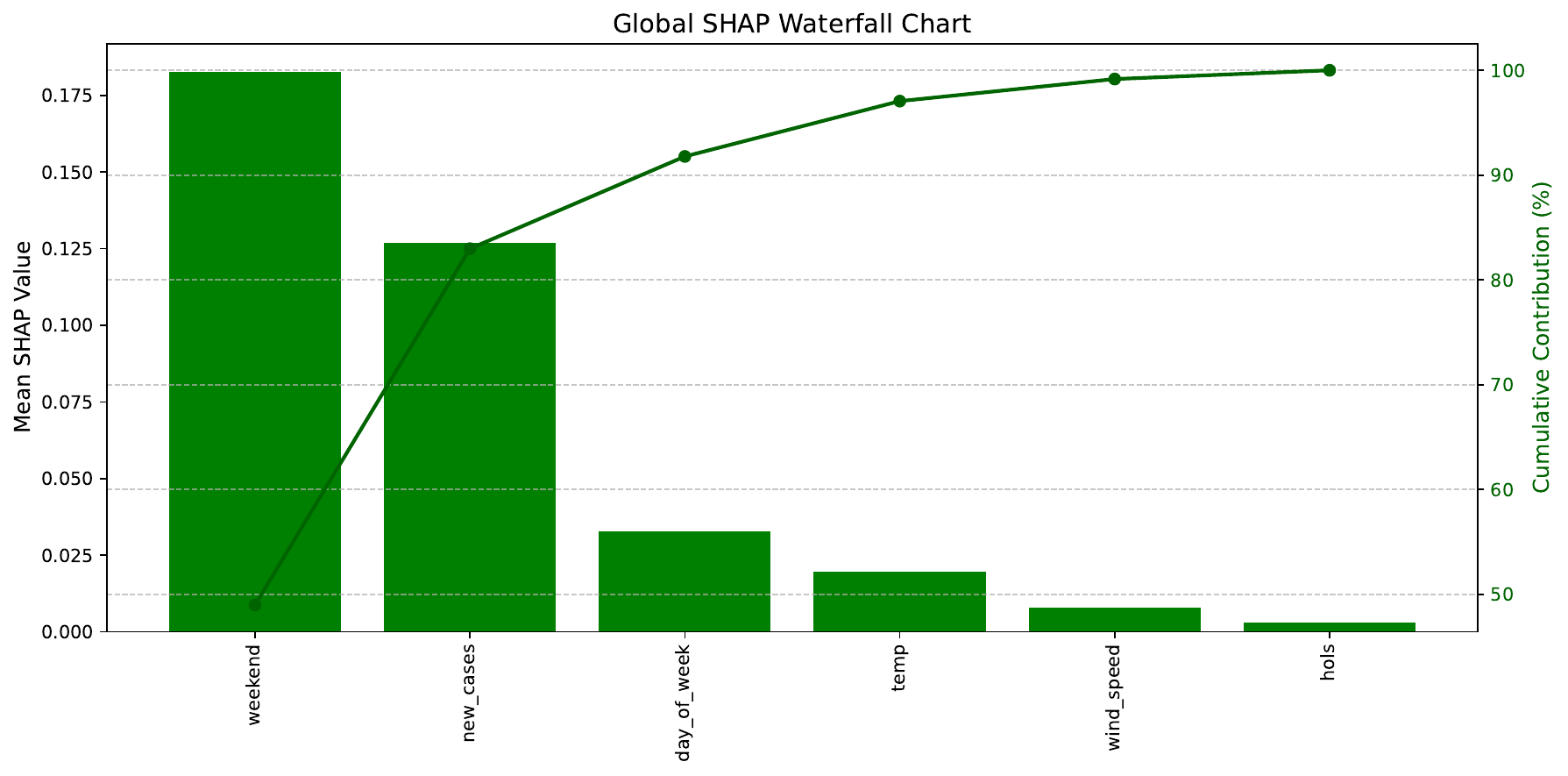}
    }
    \caption{(a) SHAP summary plot for period 1, (b) SHAP summary plot for period 3, (c) SHAP waterfall lot for period 2.}
    \label{fig:SHAP-map}
\end{figure}

Unlike differentiable models such as RNN and LSTM, SVR is non-differentiable due to its loss function, which incorporates the $\epsilon$-insensitive loss and relies on support vectors rather than optimizing all data points. As a result, IG, which measures feature contributions by integrating along a gradient-based path, is not applicable. To improve the interpretability of SVR, we employ a model-agnostic method: SHapley Additive exPlanations (SHAP), proposed by \cite{lundberg2017unified} based on game theory. SHAP calculates values for each feature, representing the average change in a model's prediction when the feature is added to different feature subsets, weighted over all possible subset combinations and orderings. These SHAP values quantify the contribution of each feature to a prediction. Figures \ref{fig:SHAP-map} (a) and (b) show SHAP summary plots for periods 1 and 3, while (c) presents a waterfall plot for period 2. In each summary plot, the horizontal axis represents SHAP values, with deeper red indicating higher feature values and blue indicating lower feature values, both influencing the SVR model's outcome.   The waterfall plot for period 2 visualizes the mean SHAP values of different features, illustrating their relative importance in the model's predictions. The bar heights represent the mean absolute SHAP values, while the line plot indicates the cumulative contribution (\%) of features to the total model variance.

In period 1, the four most significant features affecting the performance of the SVR model are weekend, day of the week, dew point, and holiday, as these features exhibit high SHAP values. A high weekend value (\emph{i.e.}, 1 for weekends) has a negative SHAP value, while a low weekend value (\emph{i.e.}, 0 for weekdays) has a positive SHAP value, indicating that weekends reduce TTI. This aligns with expectations, as people tend to stay home on weekends, leading to lower TTI, whereas TTI increases on weekdays due to commuting demands. A similar trend is observed for holidays, and to avoid redundancy, we do not elaborate further. For day of the week, we encode Monday to Sunday as 0, 1, 2, 3, 4, 5, and 6, respectively. Higher values indicate weekend samples, while lower values correspond to weekdays. Most high feature values are associated with a decrease in TTI, while most low feature values correlate with an increase, reinforcing the observation that traffic congestion is higher on weekdays due to commuting demands.
Regarding the dew point, high values create humid, foggy, or rainy conditions that reduce driving speeds, necessitate cautious braking, and increase congestion, ultimately increasing TTI. In period 3, in addition to the weekend and day of the week, which also appeared in period 1, two new significant features emerge: fuel prices and new hospitalizations. {\color{blue}Counterintuitively, higher fuel prices are associated with an increased TTI. This pattern may be context-dependent and particularly relevant to the post-COVID-19 period. Several studies have documented a decline in public transit usage due to health concerns and a modal shift toward private vehicles and ride-hailing services, even in the face of rising fuel prices after pandemic \citep{google2020covid,zhang2020effects}. Another contributing factor could be the limited extent of the fuel price increase during the study period, which may not have been sufficient to prompt a noticeable reduction in private vehicle usage. In our dataset, fuel prices increased by approximately 30\%, a change that, while evident, may not be substantial enough to meaningfully influence travel behavior. This aligns with prior findings in the literature (\emph{e.g.},\citep{goodwin2004elasticities}), which suggest that short-run travel demand tends to exhibit low price elasticity, meaning that travelers often do not significantly adjust their driving behavior in response to moderate fuel price fluctuations.
While our findings are consistent with this behavioral shift, we caution that this interpretation remains suggestive, and further data (e.g., mode choice information) would be required to validate the underlying causal mechanisms.} For new hospitalizations, higher values lead to more cautious travel behavior, reduced commuting, and fewer public activities, thereby lowering congestion and decreasing TTI. Conversely, lower new hospitalizations encourage a return to normal mobility patterns, increasing traffic volume and congestion, ultimately raising TTI. In period 2, the dominant factors affecting TTI are weekends, new cases, and day of the week, with their cumulative contribution exceeding 90\%. New cases emerge as the second most influential factor, likely due to strict government restrictions, such as remote work mandates and limitations on gatherings, aimed at reducing interactions. These measures significantly impact traffic mobility, leading to observable changes in TTI.

While this study focuses on Alameda County, the proposed modeling framework, ML prediction methods, interpretability tools, and derived insights developed here have broader applicability to other regions with similar urban and traffic characteristics. The modeling framework and prediction techniques presented in this work are designed to be generalizable, allowing for adaptation to different geographic contexts. For instance, the integration of pandemic-related policy impacts and traffic dynamics can be tailored to reflect local conditions, such as population density, transportation infrastructure, and regional COVID-19 response strategies. Additionally, the insights gained from Alameda County—such as the relationship between policy changes and traffic congestion patterns—can serve as a valuable reference for analyzing other high-density urban areas or regions experiencing similar disruptions \citep{zhang2024analysis}.
{\color{blue}
\begin{remark}
While SHAP and IG are widely used for model interpretability, both methods have known limitations. IG may suffer from noise accumulation and instability, and is particularly sensitive to the choice of baseline input. To address this issue, we tested multiple baselines in our analysis and found the attribution results to be qualitatively consistent, suggesting robustness. For models such as SVR, which are non-differentiable and therefore incompatible with IG, we applied KernelSHAP--a model-agnostic variant of SHAP. However, KernelSHAP relies on the assumption of feature independence when estimating Shapley values, which may be violated in our dataset due to correlated weather variables and temporal dependencies. Although this limitation may affect the precision of the attributions, SHAP remains one of the most widely used tools for interpreting black-box models. In this study, we treat both SHAP- and IG-based explanations as qualitative insights, and use them primarily to support broader model insights and hypothesis generation.
\end{remark}}

\subsection{Model Validation}
\begin{figure}[ht]
    \centering
    \includegraphics[width=0.9\linewidth]{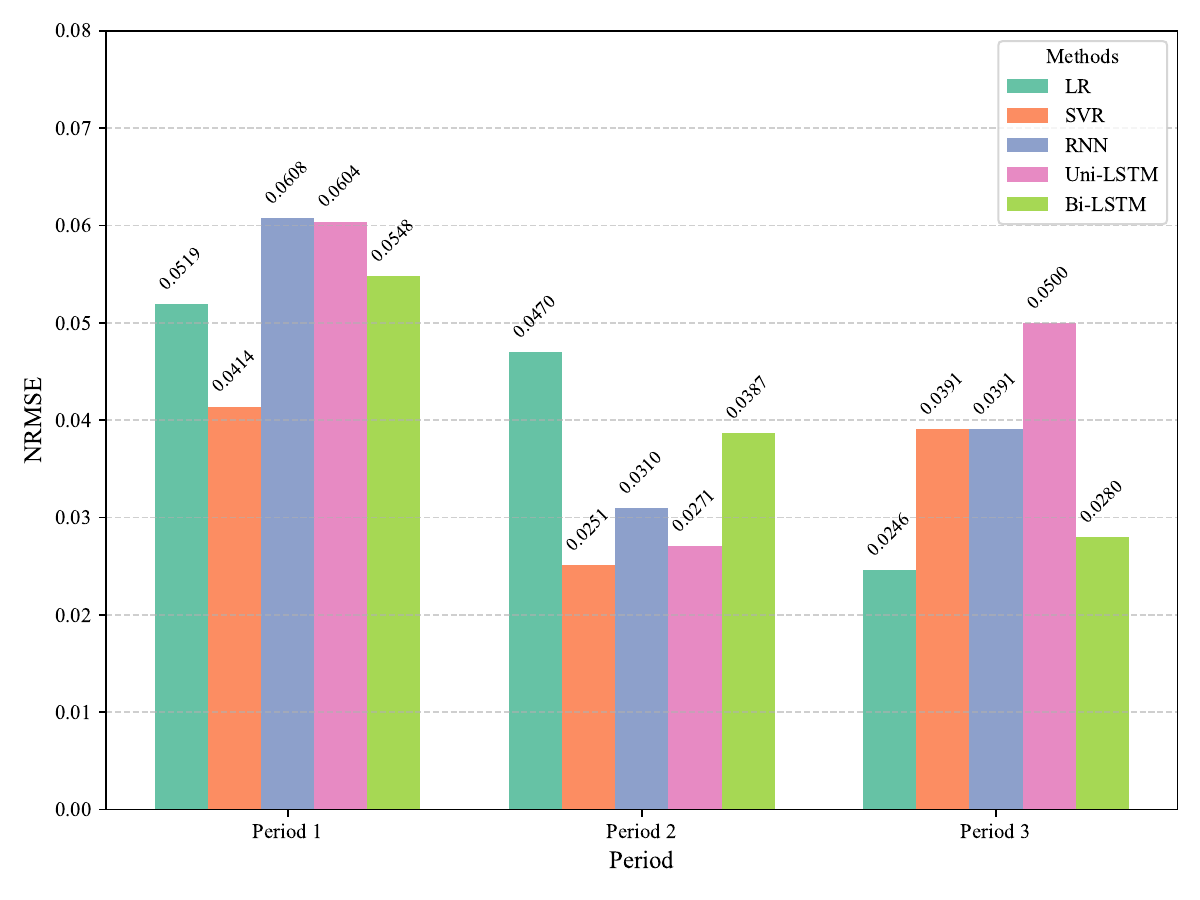}
    \caption{Comparison of ML models based on NRMSE across time periods after removing significant features}
    \label{fig:per_total_com}
\end{figure}

Note that, prior to model training, we apply RFECV to identify and retain the most relevant features for each period. We conduct RFECV separately for each pandemic phase to identify the subset of features that most significantly influence model performance. Starting with 14 candidate features, the final selections yield 9 features for Period 1 (pre-COVID), 6 for Period 2 (during COVID), and 11 for Period 3 (post-COVID). This tailored selection process ensures that each model is trained on the most relevant variables, thereby reducing the risk of overfitting and enhancing generalization across different periods.  To further validate the importance of the selected features on the prediction results, we conduct ablation tests by removing one key feature from each period based on its high importance score derived from SHAP or IG. Specifically, we remove ``weekend'' in Period 1, ``new cases'' in Period 2, and ``new hospitalizations'' in Period 3. We then retrain the model using the remaining features. Figure~\ref{fig:per_total_com} presents the NRMSE of four machine learning models after removing these important features. Compared to the original results shown in Figure~\ref{fig:per_total}, a substantial decline in prediction accuracy is observed. This provides strong empirical evidence that the features identified by RFECV are indeed critical to the models' predictive performance. 


\subsection{Discussion on Spatial Extension}
Note that the primary reason for not incorporating a finer spatial granularity (\emph{e.g.}, traffic analysis zone or census tract level) is the unavailability of certain key input features--most notably, COVID-19-related factors such as daily new cases and hospitalizations-at the sub-county level. While our target variable (the daily TTI) could in principle be modeled at a finer spatial scale, aligning all input features to such granularity was not feasible due to data constraints. As a result, all variables were aggregated at the county level to ensure consistency. Despite this limitation, our selected models are general, robust, and readily extendable to spatially disaggregated data. When zone-level input features become available, the models can be adapted as follows:

SVR does not inherently model temporal or spatial dependencies. However, spatial features can be incorporated by concatenating inputs from multiple zones into a single high-dimensional feature vector. While feasible, this approach becomes computationally intensive and lacks scalability as the number of zones increases. Similarly, MLR can be extended to incorporate spatial granularity by including zone-level features as part of a unified input vector for each time step. Spatial heterogeneity can be captured using fixed effects or zone-specific dummy variables. Although MLR does not explicitly model spatial or temporal dependencies, its simplicity enables efficient estimation and clear interpretability. To account for spatial interactions, interaction terms or spatial lag variables can also be introduced. Instead, RNN, Uni-LSTM, and Bi-LSTM models all share the capability of processing multivariate time series inputs. When spatial data becomes available, each time step's input can include concatenated features from multiple zones, resulting in an input tensor of shape \textit{(samples, time\_steps, total\_features)}, where total\_features equals the number of zones multiplied by the number of features per zone. This allows the models to learn temporal dynamics while implicitly capturing spatial interactions. While the general strategy of integrating spatial features is similar in these deep learning models, each has distinct strengths. Specifically, RNN captures sequential patterns but may suffer from vanishing gradients over longer sequences. Uni-LSTM adds gating mechanisms and memory cells to selectively retain relevant temporal and spatial information. Bi-LSTM further enhances modeling power by processing the sequence in both forward and backward directions, which is particularly beneficial when the full-sequence context is valuable.
\section{Conclusions}
\label{conclu}
In conclusion, this study effectively applied three advanced machine learning models, including SVR,  MLR,  RNN, and LSTM—to predict traffic congestion in a business district in California during the pre-lockdown, lockdown, and post-lockdown periods of the COVID-19 pandemic. We identified key external features, including weather, seasonality, and COVID-19-related factors, and then used ERFCV to prevent overfitting. Based on these selected features, we trained and optimized the models through extensive experimentation and parameter tuning.  Manual parameter tuning was used for SVR and RNN, while an adaptive parameter selection approach was applied to LSTM due to its greater number of hyperparameters and higher sensitivity to tuning compared to the other two models. By evaluating the models' performance using NRMSE, we found that Bi-LSTM performed the best across all periods. To overcome the limited interpretability of ML methods, we employed the IG method in the bi-LSTM model, which revealed that new COVID-19 cases had a significant influence on traffic patterns during the lockdown and post-lockdown periods, reflecting the significant importance of the pandemic's impact on predicting traffic congestion. Besides, three baselines—zero, mean, and past-week average—are utilized to evaluate the robustness of IG attributions. The resulting attribution maps across these baselines exhibit similar spatial and temporal patterns, suggesting the reliability of our IG-based interpretation. We also applied SHAP to SVR and observed that, from the onset of lockdowns, people became increasingly cautious. As COVID-19 severity worsened (\emph{e.g.}, the increase in new hospitalization), they preferred to stay home and opted for private vehicles or ride-hailing services over public transportation to minimize social interactions.

This study provides valuable insights for enhancing traffic management, transportation policy, and urban planning, fostering a more resilient transportation system capable of navigating future pandemic-like events. Firstly, the demonstrated strong correlation between new hospitalizations and traffic congestion underscores the necessity of integrating public health indicators into real-time traffic monitoring and prediction systems. Enhanced predictive capabilities would enable proactive traffic management strategies, optimizing transportation operations during public health crises. For instance, traffic signal timing could be dynamically adjusted to optimize traffic flow, strategically scheduling roadworks in periods of low congestion to minimize disruptions, and adapting public transit schedules and congestion pricing to fluctuating travel demands. Secondly, the prominence of new COVID-19 cases in Period 2 highlights the profound impact of government-imposed mobility restrictions on traffic patterns, emphasizing the need for adaptive traffic prediction and management systems. Granular, corridor-specific predictions would empower authorities to optimize road capacity through dynamic traffic control adjustments, such as reversible road/lane configurations, prioritizing access to critical facilities like hospitals. The modeling techniques presented herein can be applied to individual corridors, facilitating the development of targeted prediction capabilities that respond to sudden traffic volume and pattern shifts during dynamic pandemic scenarios. Furthermore, these findings highlight the necessity for planners to strategically locate social amenities and facilities, alongside designing adaptable road systems that facilitate the implementation of the aforementioned traffic management strategies during dynamic pandemic situations. Thirdly, the counterintuitive relationship between elevated fuel prices and increased congestion in Period 3 suggests persistent underutilization of public transit, despite rising driving costs. This necessitates robust incentives to promote public transit adoption, including measures to enhance safety and mitigate disease transmission, improve service reliability, and expand multimodal integration. Furthermore, investments in non-motorized infrastructure, such as dedicated cycling lanes and pedestrian-friendly urban design, can provide viable alternatives, strengthening the resilience of the transportation system in the face of evolving conditions, both predictable and unpredictable.

Future extensions of this study could involve expanding the current model to other countries, allowing for a broader evaluation of its generality and applicability in different regions. Incorporating spatiotemporal traffic dynamics across various cities/regions/corridors. would provide deeper insights into traffic patterns. Additionally, including more features, such as human mobility-based data, could further enhance the model's prediction accuracy, particularly in understanding traffic congestion during COVID-19 and similar crises.

\section{Acknowledgement}
\label{Aknlgmt}
The work was supported by Singapore Ministry of Education Academic Research Fund Tier 1 (A-8001173-00-00).

\bibliographystyle{ormsv080}
\bibliography{cite}
\appendix
\section*{Appendix A: Computation of Daily TTI in PeMS}
\addcontentsline{toc}{section}{Appendix A: Computation of Daily TTI in PeMS}

The Travel Time Index (TTI) used in this study is obtained directly from the PeMS database, where it is provided as a daily metric for each county. Although we do not compute the TTI ourselves, a brief summary of its calculation methodology in PeMS is provided below for clarification. The calculation process involves three main steps: (1) estimating travel time at the segment level, (2) aggregating across segments to the county level using traffic weights, and (3) computing the daily average. The details are as follows: 
\begin{itemize}
    \item \textbf{Segment-Level Travel Time} 
    
    For each freeway segment, PeMS estimates the \emph{actual travel time} at each 5-minute interval based on the aggregate speed (\emph{i.e.}, flow-weighted average speed across all lanes). The actual travel time is computed as:
\[
\text{TravelTime}_{i,t} = \frac{L_i}{v_{i,t}}
\]
where $L_i$ is the length of segment $i$, $v_{i,t}$ is the aggregate speed for segment $i$ at time interval $t$.

The \emph{free-flow travel time} is defined using a benchmark speed of 60 mph:
\[
\text{FreeFlowTime}_{i} = \frac{L_i}{60}
\]

The Travel Time Index (TTI) at segment $i$ and time $t$ is then calculated as:
\[
\text{TTI}_{i,t} = \frac{\text{TravelTime}_{i,t}}{\text{FreeFlowTime}_{i}} = \frac{L_i / v_{i,t}}{L_i / 60} = \frac{60}{v_{i,t}}
\]
\item \textbf{County-Level Aggregation}

PeMS aggregates segment-level TTI to the county level using Vehicle Miles Traveled (VMT) as weights. For each segment and time interval:
\[
\text{VMT}_{i,t} = q_{i,t} \cdot L_i
\]
where $q_{i,t}$ is the observed traffic flow (number of vehicles) on segment $i$ during time $t$.

The county-level TTI at time $t$ is:
\[
\text{TTI}_{\text{county}, t} = \frac{\sum_i \text{TTI}_{i,t} \cdot \text{VMT}_{i,t}}{\sum_i \text{VMT}_{i,t}}
\] 
\item \textbf{Daily Aggregation}

To obtain the daily TTI for the county, PeMS takes the arithmetic average over 288 five-minute intervals in a day:
\[
\text{TTI}_{\text{county, daily}} = \frac{1}{288} \sum_{t=1}^{288} \text{TTI}_{\text{county}, t}
\]

This daily value represents the average traffic congestion level across all freeway segments in the county over the entire day.
\end{itemize}
\end{document}